\documentclass[twocolumn]{onethree}

\usepackage[toc,page,header]{appendix}
\usepackage{minitoc}
\usepackage{amsmath,amssymb}
\usepackage{bm}
\usepackage{wrapfig}
\usepackage{enumitem}

\usepackage{graphicx}
\usepackage{booktabs}
\usepackage{multirow}
\usepackage[table]{xcolor}
\definecolor{altrow}{HTML}{EEF9FB}

\usepackage[table,dvipsnames]{xcolor} 
\usepackage[most]{tcolorbox}
\usepackage{colortbl}     
\definecolor{lightgray}{gray}{0.96} 
\usepackage{tabularx}
\usepackage{booktabs}
\usepackage{longtable}

\usepackage{arydshln}

\newcommand{\ModelIcon}[1]{%
  \raisebox{-.24ex}{%
    \includegraphics[height=1.55ex]{icon/model-family-icons/#1}%
  }\hspace{.38em}%
}

\newcommand{\IconModel}[2]{\csname #1Icon\endcsname\hspace{.38em}#2}

\definecolor{lightgray}{gray}{0.96} 
\definecolor{oursblue}{RGB}{232,240,250}
\definecolor{findingpink}{RGB}{254,225,221}
\definecolor{findingpinklight}{RGB}{254,244,243}
\definecolor{findingpinkframe}{RGB}{238,188,181}

\title{Evaluating Latent Knowledge Stability in Multimodal Large Language Models}

\author[1,4]{Chuancheng Shi*}
\author[2,4]{Shangze Li*}
\author[1]{Wenhua Wu}
\author[3]{Zifeng Cheng}
\author[4]{Pengyang Shao}
\author[4]{Fei Shen$^{\dagger}$}
\author[4]{Tat-Seng Chua}

\affiliation[1]{University of Sydney}
\affiliation[2]{Nanjing University of Science and Technology}
\affiliation[3]{Nanjing University}
\affiliation[4]{National University of Singapore, NExT++ Research Centre}

\abstract{
Changes in image viewpoint and question formulation can alter latent knowledge representations in multimodal large language models (MLLMs).
However, output correctness and overall representational drift provide limited insight into how these changes affect model decisions.
We therefore evaluate latent knowledge stability across visual and linguistic elicitation conditions from two complementary perspectives: answer preference and internal support. 
Specifically, we devise the margin retention score (MRS), which measures the first-order projected degradation of the initial answer's logit margin over competing options across conditions and assesses how changes in latent knowledge representations affect this margin. 
We also present latent knowledge support-structure stability (KSS), which evaluates reference-correct items by comparing internal support profiles constructed using the margin gradient under the reference condition, which jointly reflects changes in how internal support is distributed and changes in total support magnitude.
We further construct MMStableBench, comprising 19,452 knowledge items (facts), each probing the same latent knowledge across varied image viewpoints, semantically equivalent question formulations, and answer-option permutations. 
Experiments across 15 MLLMs show that MRS and KSS reveal latent knowledge instability not captured by accuracy, while an MRS-inspired regularizer further improves the generalization of edited latent knowledge.
\vspace{-0.1cm}
}

\checkdata[Resources]{
  \href{https://github.com/LucasShi1101/MLLM_Stable_Valid}{\raisebox{-1ex}{\includegraphics[height=4.2ex]{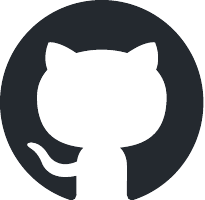}}~Code}\hspace{1em}
  \href{https://huggingface.co/datasets/LucasLoading/MMStableBench}{\raisebox{-1ex}{\includegraphics[height=4.2ex]{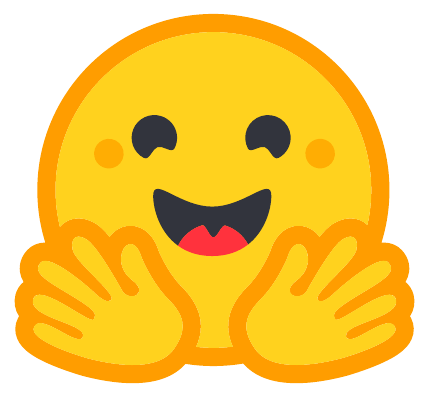}}~Dataset}\hspace{1em}
  \href{mailto:cshi0459@uni.sydney.edu.au}{\nolinkurl{cshi0459@uni.sydney.edu.au}}\hspace{1em}
  \href{mailto:shangzeli@njust.edu.cn}{\nolinkurl{shangzeli@njust.edu.cn}}
  \vspace{0.2cm}
}

\checkdata[Date]{
  November 24, 2026
}

\begin{document}
\maketitle

\begingroup
  \renewcommand{\thefootnote}{\fnsymbol{footnote}}
  \footnotetext[1]{Equal Contribution.}
  \footnotetext[2]{Corresponding Author (\href{mailto:shenfei29@nus.edu.sg}{shenfei29@nus.edu.sg}).}
\endgroup

\section{Introduction}
Multimodal large language models (MLLMs)~\cite{zhang2025cross,wang2025towards,khayatkhoei2025mllms,qwen3} have demonstrated strong capabilities in visual question answering and knowledge-intensive reasoning~\cite{yue2024mmmu} by integrating visual information with internal knowledge. However, changes in image viewpoints and question formulations can alter the internal representations of the same knowledge item, even when the model produces correct answers. Existing evaluations~\cite{liu2024mmbench,chen2024mmstar} primarily focus on output correctness, while overall representational drift alone provides limited insight into how such changes affect answer preference and internal knowledge support. Consequently, whether MLLMs maintain stable latent knowledge across different visual~\cite{zhang2024unveiling} and linguistic~\cite{chou2025mm} elicitation conditions remains insufficiently understood.

\begin{figure}[t]
\centering
\includegraphics[width=0.98\linewidth]{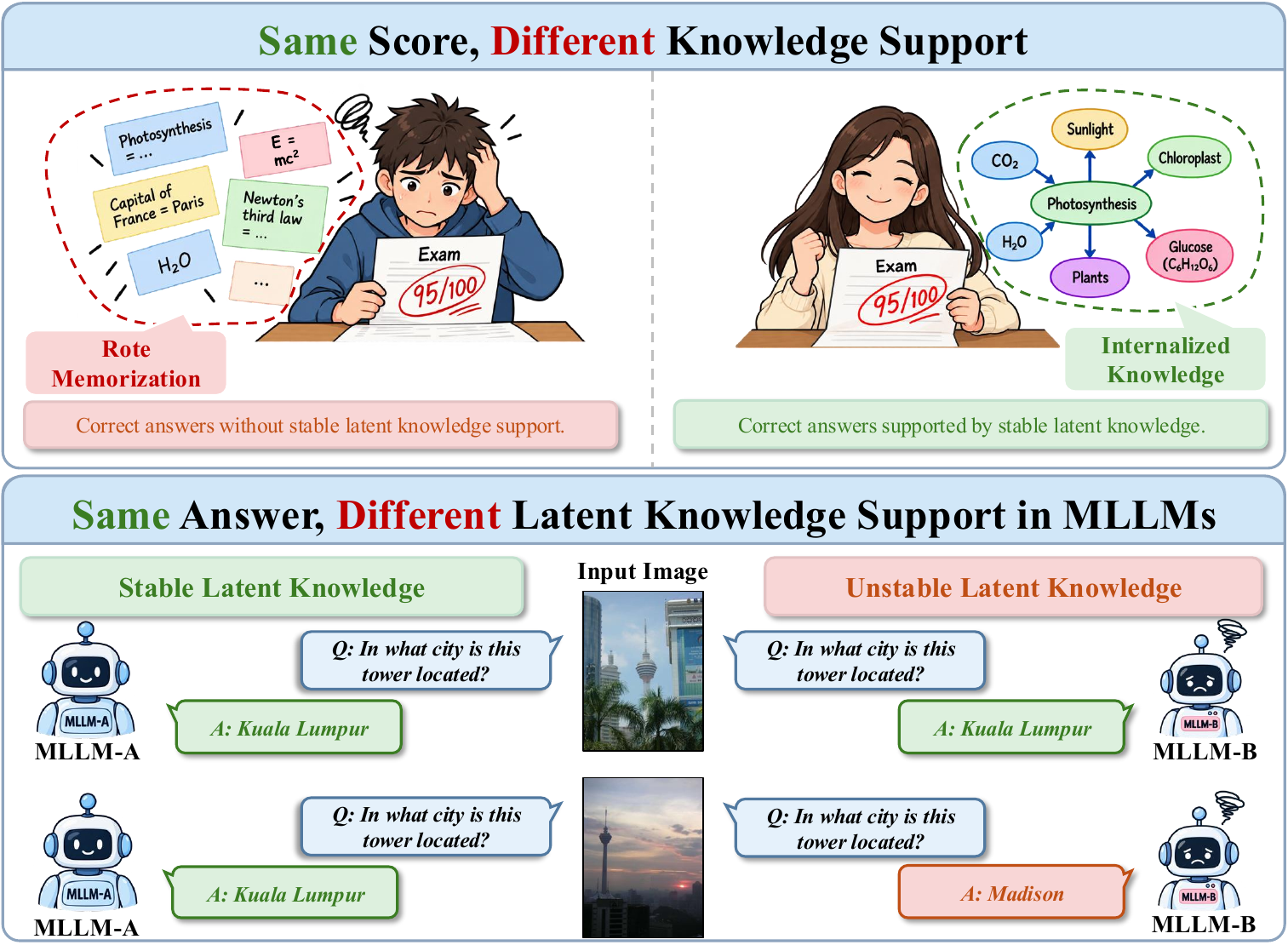}
\caption{\textbf{Same Answer, Different Latent Knowledge Stability.} (Top) Identical exam scores may reflect different levels of knowledge mastery. (Bottom) MLLMs may initially give the same correct answer but exhibit different stability across visual conditions.}
\label{fig:introduction}
\vspace{-0.6cm}
\end{figure}

Similar observable performance~\cite{Deng_2025_CVPR,parcalabescu-frank-2023-mm,gekhman2025inside} does not necessarily imply the same level of knowledge mastery. For example, students may achieve identical exam scores through rote memorization or deep understanding. While the former often relies on familiar question formats, the latter enables more flexible application of acquired knowledge (Figure~\ref{fig:introduction}, top). A similar phenomenon may arise in MLLMs. Although different models may initially provide the same correct answer, their responses can diverge when image viewpoints or question formulations~\cite{lee2026language,jegham2025visual,zhang2024unveiling,chou2026test} change. Some models continue to answer correctly, while others fail to recognize previously demonstrated knowledge (Figure~\ref{fig:introduction}, bottom). This observation motivates the correctness–stability hypothesis: for a reference-correct knowledge item, greater stability of its internal representations is associated with a higher likelihood of maintaining correctness across other elicitation conditions. Evaluating this relationship requires examining not only observable outputs but also the internal representations supporting them.

To investigate this hypothesis, we systematically evaluate latent knowledge stability in MLLMs from two complementary perspectives: answer preference and internal knowledge support. Specifically, we introduce two metrics: margin retention score (MRS) and latent knowledge support-structure stability (KSS). MRS measures whether changes in elicitation conditions degrade the logit margin between the initial answer and competing options, and examines how changes in latent knowledge representations affect this margin. KSS evaluates reference-correct knowledge items by comparing internal support profiles constructed using the reference-condition margin gradient, which jointly reflects changes in how internal support is distributed and how much support remains. Together, these metrics characterize how multimodal variations affect both the model's preference for its initial answer and the internal knowledge support underlying correct answers.

To enable systematic evaluation, we construct MMStableBench, a benchmark comprising 19,452 knowledge items with controlled image viewpoints, semantically equivalent question formulations, and answer-option permutations. We evaluate 15 MLLMs across four model families and investigate how latent knowledge stability relates to output correctness and model scaling. Our experiments reveal that models with similar output performance can exhibit different levels of internal stability, while MRS and KSS capture complementary effects of visual and linguistic elicitation conditions. Beyond evaluation, we incorporate an MRS-inspired regularizer into existing multimodal knowledge editing methods, improving the generalization of edited latent knowledge across different images of the same entity. Our contributions are summarized as follows:
\begin{itemize}
\item We investigate the correctness–stability hypothesis in MLLMs, examining whether the latent knowledge supporting reference-correct items remains stable across visual and linguistic elicitation conditions.
\item We introduce two complementary metrics, MRS and KSS, to evaluate latent knowledge stability through the robustness of answer preferences and the consistency of internal knowledge support.
\item We construct MMStableBench, comprising 19,452 knowledge items with controlled visual and linguistic variations, enabling systematic evaluation of latent knowledge stability across diverse MLLMs.
\item We demonstrate the practical utility of an MRS-inspired regularizer in multimodal knowledge editing, improving the generalization of edited latent knowledge across different images of the same entity.

\end{itemize}

\section{Related Work}

\noindent\textbf{Evaluation of MLLMs.}
Existing evaluations~\cite{fu2026mme,li2023seed} of multimodal large language models (MLLMs) primarily focus on observable performance, with benchmarks such as MMBench~\cite{liu2024mmbench}, MMMU~\cite{yue2024mmmu}, and MMStar~\cite{chen2024mmstar} assessing multimodal perception, reasoning, and knowledge-intensive capabilities. Recent studies have further investigated model robustness and consistency~\cite{chen2025mvi,jegham2025visual,weialpha2026multimodal,sterz2025dare}. MM-R³~\cite{chou2025mm} evaluates response consistency under question rephrasing and image restyling, while V²R-Bench~\cite{fan2025v2bench} examines robustness to visual variations and analyzes potential sources of failure. However, these studies primarily characterize behavioral robustness rather than the stability of latent knowledge supporting correct answers. In contrast, MMStableBench evaluates the same knowledge item across controlled visual and linguistic elicitation conditions, while MRS and KSS measure the robustness of prediction preferences and the consistency of internal knowledge support, respectively. This allows us to investigate whether consistent model outputs are accompanied by stable latent knowledge representations, providing an internal perspective complementary to existing MLLM evaluations.

\noindent\textbf{Latent Knowledge in MLLMs.}
Recent studies~\cite{deng2026causal,wang2025towards,zhang2025cross,lietzow2026vision} have revealed that pretrained multimodal models encode latent knowledge within their internal representations. LAKE~\cite{li2026latent} identifies anomaly-sensitive neurons to elicit intrinsic anomaly knowledge, while DNA~\cite{dou2026dna} uncovers discriminative neural units associated with forgery detection. SafeNexus~\cite{yu2026safenexus} further demonstrates the functional importance of internal neurons by showing that suppressing safety-related neurons degrades model performance, whereas amplifying their activations enhances safety behavior. Similarly, TraceRouter~\cite{shi2026tracerouter} and TraceRouter++~\cite{shi2026tracerouter++} identify and intervene in distributed causal pathways responsible for harmful semantic propagation. Collectively, these studies provide evidence that internal neurons~\cite{fang2025can} and their interactions contribute to supporting model capabilities. However, while prior work primarily focuses on identifying and manipulating such internal components, whether their support remains stable across different elicitation conditions is less understood. We address this question by systematically evaluating the consistency of internal knowledge support across visual and linguistic variations.

\section{Method}
\label{sec:method}

We evaluate latent knowledge stability in MLLMs across visual and linguistic elicitation conditions using two complementary metrics, margin retention score (MRS) and latent knowledge support-structure stability (KSS). MRS measures whether changes in elicitation conditions weaken the model's preference for its reference prediction, while KSS evaluates whether reference-correct items retain consistent internal knowledge support across conditions.

\subsection{Margin Retention Score}
\label{sec:mrs}

Let $i$ denote a multiple-choice knowledge item with $K_i$ answer options. We denote its reference input as $x_{i,a}$ and its target input as $x_{i,t}$, where $t\in\mathcal T_i$ indexes alternative elicitation conditions. These conditions vary image viewpoints, question formulations, or answer-option order while preserving the underlying knowledge. For model $m$ and condition $c\in\{a,t\}$, let $\boldsymbol z_{m,i,c}\in\mathbb R^{K_i}$ denote the answer-choice logits and $\boldsymbol h_{m,i,c}^{\ell}$ the final-prompt-token residual state at decoder layer $\ell$. Answer logits are aligned by semantic choice identity across conditions. Let $\mathcal L_m$ denote the set of evaluated decoder layers.

MRS measures whether changes in elicitation conditions reduce the model's preference for its reference prediction. Let $\hat y_{m,i}=\arg\max_k z_{m,i,a,k}$ denote the reference prediction, where $k$ indexes candidate answers. We define its floored logit margin over each competing answer $k$ as
\begin{equation}
g_{m,i,k}
=\max\!\left\{
z_{m,i,a,\hat y_{m,i}}-z_{m,i,a,k},\,\tau
\right\},
\qquad k\ne\hat y_{m,i},
\label{eq:mrs-anchor-gap}
\end{equation}
where $\tau>0$ is a numerical floor. For a logit change vector $\boldsymbol\delta\in\mathbb R^{K_i}$, we define the normalized margin degradation as
\begin{equation}
\mathcal D_{m,i}(\boldsymbol\delta)
=\max_{k\ne\hat y_{m,i}}
\left[
\frac{\delta_k-\delta_{\hat y_{m,i}}}
{g_{m,i,k}}
\right]_+,
\qquad [v]_+=\max(v,0).
\label{eq:mrs-damage}
\end{equation}
Here, $\delta_k$ denotes the logit change of option $k$. The positive-part operator ensures that only reductions in the reference prediction's relative advantage are penalized. MRS does not require the reference prediction to be correct.

To examine how internal representational changes affect this margin, we define the cross-condition displacement $\Delta\boldsymbol h_{m,i,t}^{\ell}=\boldsymbol h_{m,i,t}^{\ell}-\boldsymbol h_{m,i,a}^{\ell}$ and propagate it to the answer logits through the reference logit Jacobian~\cite{kramar2024atp},
\begin{equation}
\boldsymbol\delta_{m,i,t}^{\ell}=J_{m,i,a}^{\ell}\,\Delta\boldsymbol h_{m,i,t}^{\ell},
\qquad
J_{m,i,a}^{\ell}=\left.\frac{\partial\boldsymbol z_{m,i}}{\partial\boldsymbol h_{m,i}^{\ell}}\right|_{x_{i,a}}.
\label{eq:mrs-change}
\end{equation}
MRS then averages the normalized margin degradation over the evaluated layers,
\begin{equation}
\operatorname{MRS}_{m,i,t}
=\left(
1+\frac{1}{|\mathcal L_m|}
\sum_{\ell\in\mathcal L_m}
\mathcal D_{m,i}\!\left(\boldsymbol\delta_{m,i,t}^{\ell}\right)
\right)^{-1}.
\label{eq:mrs-score}
\end{equation}
Higher MRS indicates greater resistance to margin degradation.

\subsection{Latent Knowledge Support-Structure Stability}
\label{sec:kss}

KSS measures whether reference-correct items retain consistent internal knowledge support across visual and linguistic elicitation conditions. It is evaluated when $\hat y_{m,i}=y_i$, where $y_i$ denotes the verified answer. Let $r_{m,i}=\arg\max_{k\ne y_i}z_{m,i,a,k}$ denote the strongest competing answer under the reference condition. At each evaluated layer $\ell\in\mathcal L_m$, we compute the reference gradient
\begin{equation}
\boldsymbol\gamma_{m,i}^{\ell}
=\nabla_{\boldsymbol h_{m,i,a}^{\ell}}
\left(z_{m,i,a,y_i}-z_{m,i,a,r_{m,i}}\right).
\label{eq:kssl-gradient}
\end{equation}
The gradient $\boldsymbol\gamma_{m,i}^{\ell}$ measures the local sensitivity of the verified-answer margin to the hidden state and is held fixed across all elicitation conditions of knowledge item $i$.

For simplicity, we omit the indices $(m,i)$ below. We partition the hidden dimensions at layer $\ell$ into channel blocks $\mathcal B_{\ell,b}$, where $b$ indexes the blocks. For input $x$, we define the nonnegative reference-gradient projection and its total support mass as
\begin{align}
d_b^{\ell}(x)
&=\left[
\sum_{q\in\mathcal B_{\ell,b}}
\gamma_q^{\ell}h_q^{\ell}(x)
\right]_+,
\label{eq:kssl-support}\\
M^{\ell}(c)
&=\sum_b d_b^{\ell}(x_{i,c}),
\qquad c\in\{a,t\}.
\label{eq:kssl-total-mass}
\end{align}
Here, $q$ indexes hidden dimensions, $\gamma_q^{\ell}$ is the corresponding gradient component, and $h_q^{\ell}(x)$ is the activation of that dimension under input $x$. $d_b^{\ell}(x)$ represents the positive projection of block $b$ onto the fixed reference gradient, while $M^{\ell}(c)$ denotes the total positive support under condition $c$.

A layer is identifiable if $\min\{M^{\ell}(a),M^{\ell}(t)\}>\epsilon$, where $\epsilon>0$ is a numerical detection threshold. At each identifiable layer, we compare the reference and target support profiles with a weighted Dice overlap,
\begin{equation}
\operatorname{KSS}^{\ell}
=\frac{2\sum_b\min\!\left\{d_b^{\ell}(x_{i,a}),\,d_b^{\ell}(x_{i,t})\right\}}
{M^{\ell}(a)+M^{\ell}(t)},
\label{eq:kssl-layer}
\end{equation}
and average over the set of identifiable layers $\mathcal L^{\mathrm{id}}\subseteq\mathcal L_m$:
\begin{equation}
\operatorname{KSS}
=\frac{1}{|\mathcal L^{\mathrm{id}}|}
\sum_{\ell\in\mathcal L^{\mathrm{id}}}
\operatorname{KSS}^{\ell}.
\label{eq:kssl-score}
\end{equation}
KSS equals one only when support profiles coincide in distribution and total magnitude: it reduces to the overlap of normalized support distributions when $M^{\ell}(a)=M^{\ell}(t)$, and to the magnitude ratio $2\min\{M^{\ell}(a),M^{\ell}(t)\}/(M^{\ell}(a)+M^{\ell}(t))$ when the normalized distributions are identical. Thus a single score captures both redistribution and changes in the total magnitude of internal support.

We additionally use knowledge-mismatched inputs to assess whether the measured support is specific to the underlying knowledge. For a different knowledge item $j\ne i$ with reference input $x_{j,a}$, we compute its support profile using item $i$'s fixed gradient $\boldsymbol\gamma_{m,i}^{\ell}$ and channel partition. This allows us to compare $d_{m,i,b}^{\ell}(x_{i,a})$ and $d_{m,i,b}^{\ell}(x_{j,a})$ without changing the reference directions.
Sensitivity analyses of channel partitioning are in the appendix.

\begin{figure}[t]
\centering
\includegraphics[width=\linewidth]{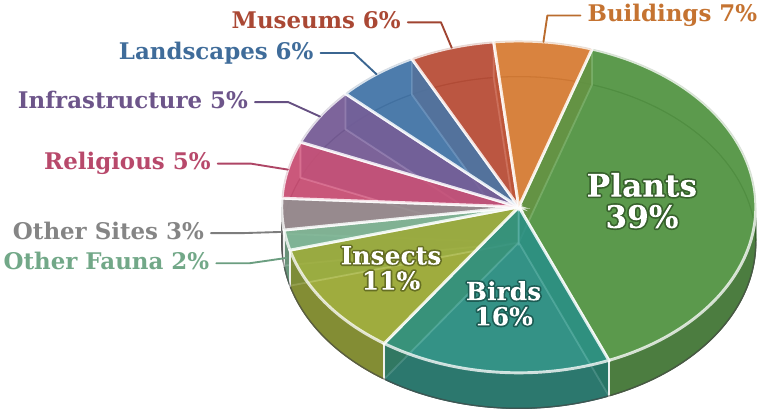}
\caption{\textbf{Category Distribution of MMStableBench.} Distribution of 19,452 knowledge items across biological species and landmark categories.}
\vspace{-0.5cm}
\label{fig:mmstable}
\end{figure}

\begin{table*}[t]
\centering
\small
\caption{\textbf{Model-Family and Scaling Comparison of Internal Stability.}
ACC, MRS, and KSS on 19{,}452 knowledge items. MRS is averaged over all knowledge items; KSS over reference-correct items. \textit{Natural} averages the closer- and farther-view images, and \textit{Translation} re-asks each item in another language. \textit{Average} is taken over the five target conditions, excluding the reference condition. Best and second-best results are shown in \textbf{bold} and \underline{underlined}.}
\vspace{-0.2cm}
\label{tab:mrs_kss_leaderboard}
\resizebox{\linewidth}{!}{%
\begin{tabular}{clccccccccccc}
\toprule
 & & & \multicolumn{2}{c}{\textit{Paraphrase}}
       & \multicolumn{2}{c}{\textit{Natural}}
       & \multicolumn{2}{c}{\textit{Opt. Perm.}}
       & \multicolumn{2}{c}{\textit{Translation}}
       & \multicolumn{2}{c}{\textit{Average}} \\
\cmidrule(lr){4-5} \cmidrule(lr){6-7} \cmidrule(lr){8-9} \cmidrule(lr){10-11} \cmidrule(lr){12-13}
Family & Model & ACC $\uparrow$
       & MRS $\uparrow$ & KSS $\uparrow$
       & MRS $\uparrow$ & KSS $\uparrow$
       & MRS $\uparrow$ & KSS $\uparrow$
       & MRS $\uparrow$ & KSS $\uparrow$
       & MRS $\uparrow$ & KSS $\uparrow$ \\
\midrule
\rowcolor{altrow}
\cellcolor{white} & \ModelIcon{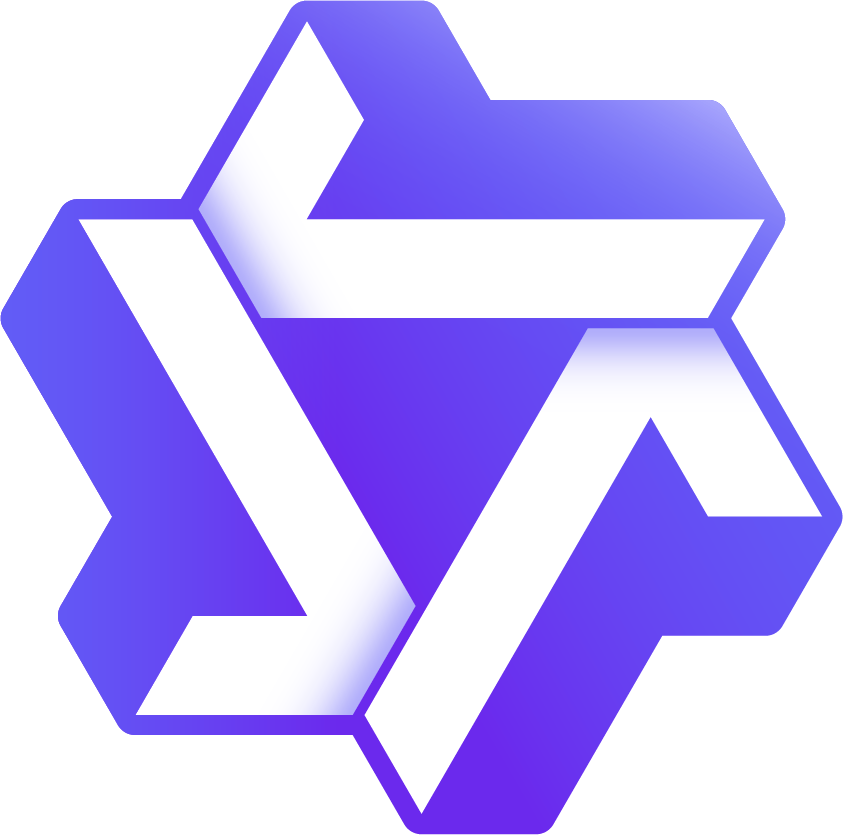}Qwen3.5-0.8B & 0.681 & 0.932 & \underline{0.952} & 0.923 & 0.937 & 0.789 & 0.929 & 0.882 & 0.800 & 0.890 & 0.911 \\
 & \ModelIcon{qwen.png}Qwen3.5-2B & 0.742 & 0.933 & 0.946 & 0.934 & \underline{0.938} & 0.849 & \underline{0.939} & 0.744 & 0.749 & 0.879 & 0.902 \\
\rowcolor{altrow}
\cellcolor{white} & \ModelIcon{qwen.png}Qwen3.5-4B & 0.838 & 0.917 & 0.920 & 0.923 & 0.907 & 0.721 & 0.851 & 0.866 & 0.756 & 0.870 & 0.868 \\
 & \ModelIcon{qwen.png}Qwen3.5-9B & 0.868 & \textbf{0.976} & 0.942 & \underline{0.970} & 0.927 & \underline{0.898} & 0.914 & 0.918 & 0.793 & \underline{0.946} & 0.901 \\
\rowcolor{altrow}
\cellcolor{white}\multirow{-5}{*}{\textit{Qwen}~\cite{qwen3}}
 & \ModelIcon{qwen.png}Qwen3.5-27B & \textbf{0.894} & 0.957 & 0.948 & 0.954 & \textbf{0.950} & 0.855 & 0.902 & 0.915 & \textbf{0.857} & 0.927 & \underline{0.921} \\
\hdashline
 & \ModelIcon{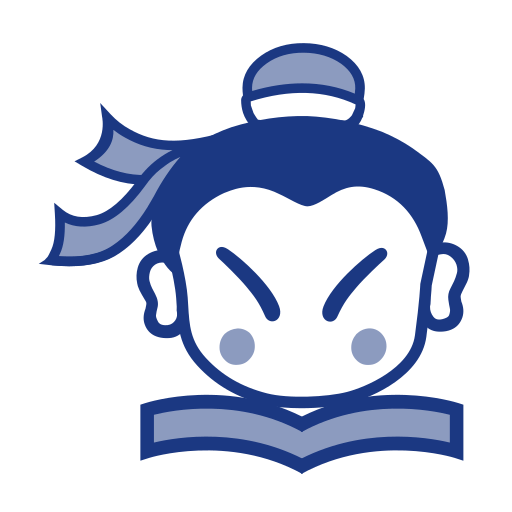}InternVL3.5-1B & 0.659 & 0.855 & \textbf{0.953} & 0.836 & 0.933 & 0.613 & 0.921 & 0.712 & 0.825 & 0.770 & 0.913 \\
\rowcolor{altrow}
\cellcolor{white} & \ModelIcon{internvl.png}InternVL3.5-2B & 0.729 & 0.867 & 0.944 & 0.862 & 0.935 & 0.647 & 0.916 & 0.832 & 0.817 & 0.814 & 0.909 \\
 & \ModelIcon{internvl.png}InternVL3.5-4B & 0.823 & 0.894 & 0.924 & 0.897 & 0.912 & 0.713 & 0.881 & 0.862 & 0.804 & 0.853 & 0.887 \\
\rowcolor{altrow}
\cellcolor{white} & \ModelIcon{internvl.png}InternVL3.5-8B & 0.842 & 0.912 & 0.935 & 0.919 & 0.933 & 0.768 & 0.905 & \underline{0.924} & 0.821 & 0.888 & 0.906 \\
 & \ModelIcon{internvl.png}InternVL3.5-14B & 0.851 & 0.906 & 0.939 & 0.920 & \underline{0.938} & 0.767 & 0.907 & 0.735 & 0.793 & 0.850 & 0.903 \\
\rowcolor{altrow}
\cellcolor{white}\multirow{-6}{*}{\textit{InternVL}~\cite{wang2025internvl3}}
 & \ModelIcon{internvl.png}InternVL3.5-38B & \underline{0.870} & \underline{0.964} & 0.947 & \textbf{0.974} & \textbf{0.950} & \textbf{0.971} & \textbf{0.954} & \textbf{0.954} & \underline{0.847} & \textbf{0.967} & \textbf{0.930} \\
\hdashline
 & \ModelIcon{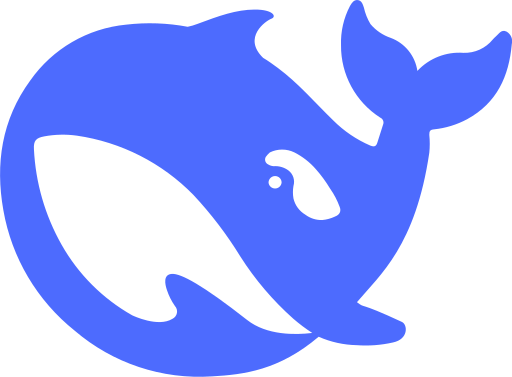}DeepSeek-VL-1.3B & 0.740 & 0.878 & 0.896 & 0.891 & 0.910 & 0.624 & 0.736 & 0.812 & 0.814 & 0.819 & 0.853 \\
\rowcolor{altrow}
\cellcolor{white}\multirow{-2}{*}{\textit{DeepSeek-VL}~\cite{lu2024deepseek}}
 & \ModelIcon{deepseek.png}DeepSeek-VL-7B & 0.771 & 0.905 & 0.861 & 0.904 & 0.877 & 0.713 & 0.778 & 0.879 & 0.795 & 0.861 & 0.838 \\
\hdashline
 & \ModelIcon{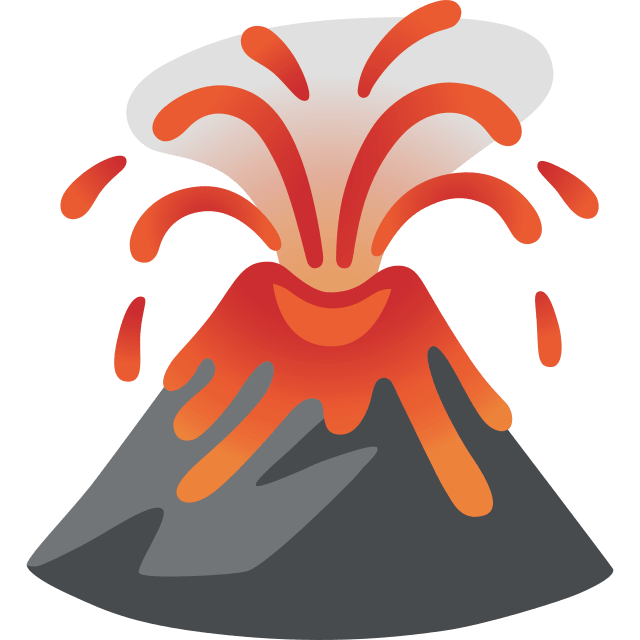}LLaVA-1.5-7B & 0.774 & 0.832 & 0.877 & 0.869 & 0.913 & 0.412 & 0.646 & 0.803 & 0.827 & 0.757 & 0.835 \\
\rowcolor{altrow}
\cellcolor{white}\multirow{-2}{*}{\textit{LLaVA}~\cite{liu2024improved}}
 & \ModelIcon{llava.png}LLaVA-1.5-13B & 0.801 & 0.834 & 0.842 & 0.863 & 0.880 & 0.452 & 0.627 & 0.793 & 0.800 & 0.761 & 0.806 \\
\bottomrule
\end{tabular}%
}
\vspace{-0.4cm}
\end{table*}

\section{Experiments and Analysis}

\noindent\textbf{Construction of MMStableBench.}
We construct MMStableBench from facts about plant, animal species and landmarks, each linked to a Wikipedia article and multiple real-world photographs of the same entity. Qwen2.5-32B-Instruct~\cite{yang2024qwen25} generates questions, including template-based, passage-grounded, multi-answer, and two-hop questions; four researchers spot-check a subset for factual accuracy and evidence consistency. We treat each entity--question--answer tuple as a fact and merge duplicate records. We use text-only probes across four model families to retain facts they know. A fact is kept only if both wrappers are answered correctly by all four families under normalized exact match or BEM~\cite{bulian2022tomayto} at a threshold of 0.5, which retains 21,355 of 205,194 candidate facts. Qwen2.5-32B-Instruct generates paraphrases, which undergo automatic validation; close- and far-view photographs are selected and checked automatically. The benchmark contains 19,452 facts (Figure~\ref{fig:mmstable}), comprising 13,420 species-related facts and 6,032 landmark-related facts. Each fact is formatted as a four-choice question under six semantically equivalent conditions, namely a reference condition and five target conditions: question paraphrase, close-view image, far-view image, answer-option permutation~\cite{rosenthal2026unexplored}, and translation.

\noindent\textbf{Protocol.}
We evaluate 15 MLLMs from four model families, including Qwen3.5~\cite{qwen3}, InternVL3.5~\cite{wang2025internvl3}, DeepSeek-VL~\cite{lu2024deepseek}, and LLaVA-1.5~\cite{liu2024improved}, on MMStableBench. All models are evaluated using the same four-choice knowledge items under controlled visual and linguistic elicitation conditions, including question paraphrasing, natural image viewpoints, answer-option permutations, and translation. Predictions are obtained from answer-choice logits aligned by semantic identity across conditions. We compute ACC from output correctness, MRS over all knowledge items, and KSS over items answered correctly under the reference condition with identifiable internal support. We use $\mathcal L_m$ = decoder layers at 50\%, 65\%, and 80\% depth and $\tau=0.5$ for all experiments. More details are in the Appendix.

\subsection{Model Rankings}

\noindent\textbf{Model Size, Accuracy, and Stability.}
To investigate whether model scaling and output correctness reflect latent knowledge stability in MLLMs, we evaluate 15 models from four model families on MMStableBench, comparing ACC, MRS, and KSS across elicitation conditions. Table~\ref{tab:mrs_kss_leaderboard} shows that higher output accuracy and larger model size do not necessarily lead to greater latent knowledge stability, with models exhibiting varying sensitivity to input changes. As InternVL3.5~\cite{wang2025internvl3} scales from 1B to 38B, ACC increases from 0.659 to 0.870, and average MRS improves from 0.770 to 0.967, whereas average KSS changes only slightly from 0.913 to 0.930, with non-monotonic variations across model sizes. Similarly, scaling Qwen3.5~\cite{qwen3} from 9B to 27B decreases average MRS from 0.946 to 0.927, while ACC increases from 0.868 to 0.894 and average KSS increases from 0.901 to 0.921. Different elicitation conditions affect model stability differently. For example, LLaVA-1.5-7B~\cite{liu2024improved} achieves MRS and KSS scores of 0.832 and 0.877 under paraphrasing, respectively, but these scores decline to 0.412 and 0.646 under option permutations. These findings suggest that model scaling and output correctness alone cannot fully characterize latent knowledge stability, while MRS and KSS reveal complementary differences in prediction robustness and internal knowledge support across elicitation conditions.

\begin{figure}[t]
\centering
\includegraphics[width=\linewidth]{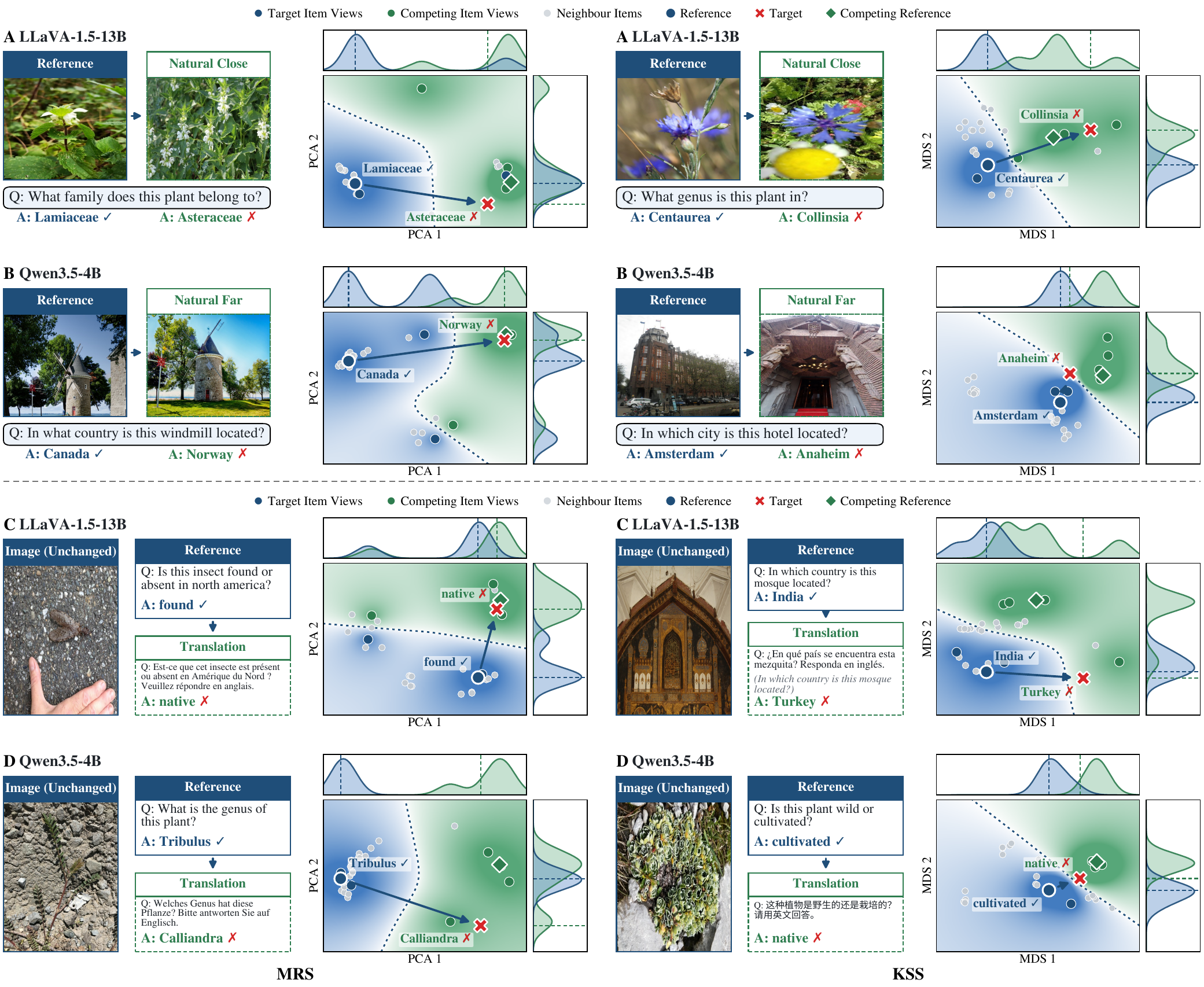}
\vspace{-0.5cm}
\caption{\textbf{Qualitative Analysis under Visual and Linguistic Variations.} Natural-view changes (top) and question translations with unchanged images (bottom) can alter model predictions.}
\vspace{-0.4cm}
\label{fig:qualitative}
\end{figure}

\begin{figure}[t]
\centering
\includegraphics[width=\linewidth]{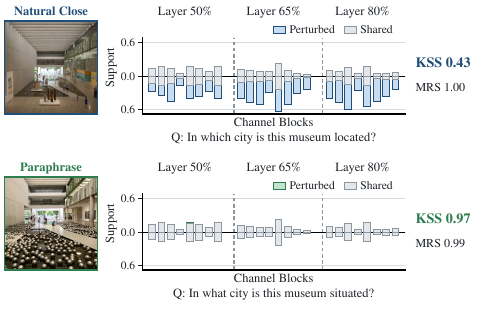}
\vspace{-0.2cm}
\caption{\textbf{Internal Support Divergence.} Both perturbed inputs preserve the correct answer with near-maximal MRS, yet exhibit substantially different KSS.}
\label{fig:fig_kss_strips}
\vspace{-0.3cm}
\end{figure}

\noindent\textbf{Qualitative Analysis under Visual Variations.}
To investigate how visual variations affect latent knowledge stability, we compare natural close- and far-view images of the same entity while keeping the question unchanged. As shown in the upper half of Figure~\ref{fig:qualitative}, LLaVA-1.5-13B~\cite{liu2024improved} changes its answer from the correct plant family Lamiaceae to Asteraceae under a natural close view, while Qwen3.5-4B~\cite{qwen3} changes its answer from Canada to Norway for a windmill under a natural far view. The MDS panels accompanying KSS present distinct examples: LLaVA-1.5-13B~\cite{liu2024improved} changes its prediction from Centaurea to Collinsia, and Qwen3.5-4B~\cite{qwen3} from Amsterdam to Anaheim.
Together, these cases illustrate that knowledge-preserving viewpoint
changes can induce answer flips and changes in projected internal
representations, motivating stability evaluation beyond output
correctness.

\begin{figure}[t]
\centering
\includegraphics[width=\linewidth]{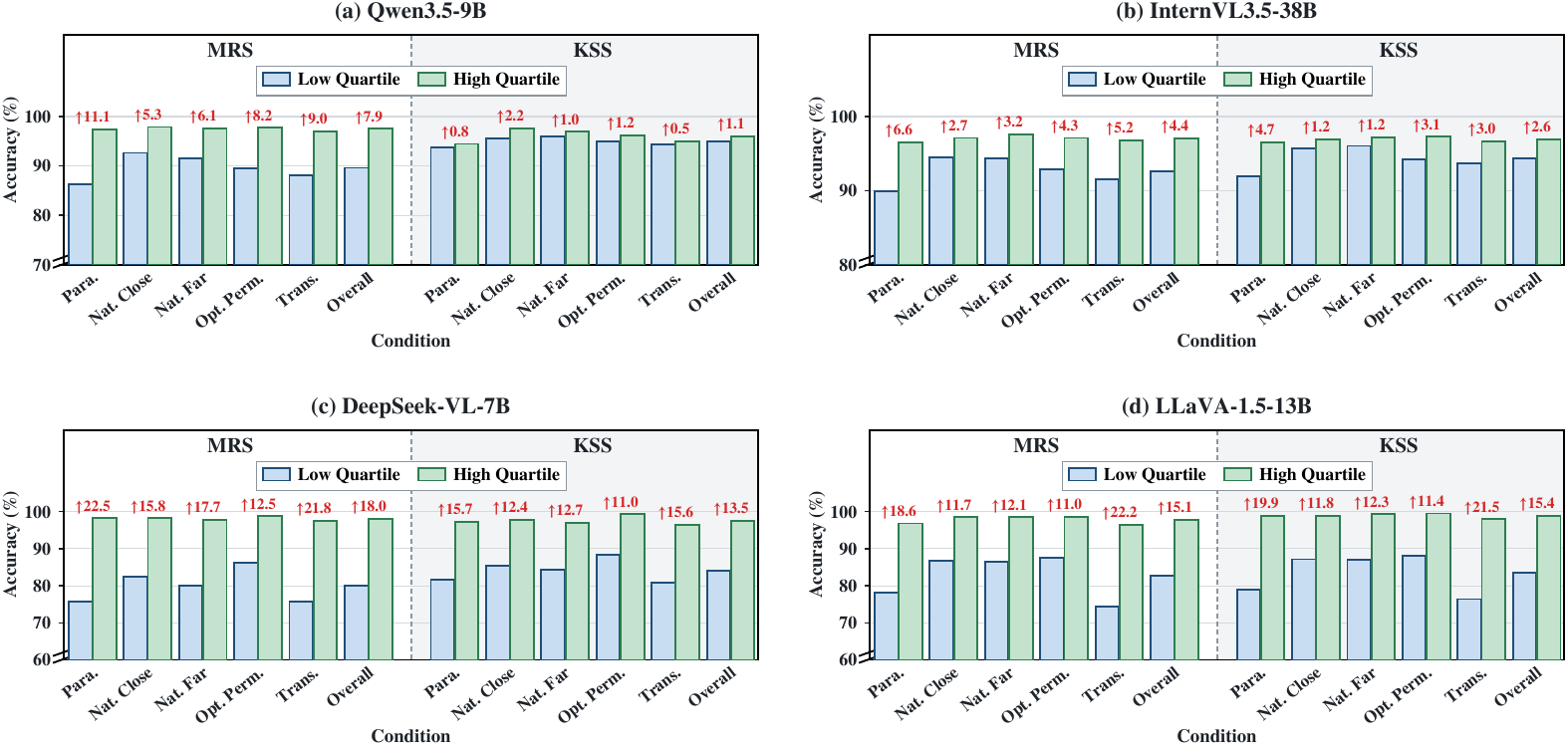}
\caption{\textbf{Latent Knowledge Stability and Output Correctness.} Accuracy on each held-out condition for high- and low-stability quartiles, with stability averaged over the remaining conditions (leave-one-condition-out), computed on items answered correctly under the reference condition.}
\label{fig:fig4}
\vspace{-0.4cm}
\end{figure}

\begin{figure}[t]
\centering
\includegraphics[width=\linewidth]{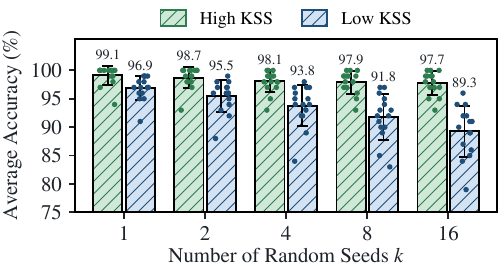}
\vspace{-0.3cm}
\caption{\textbf{Robustness Under Repeated Random Perturbations.}
Percentage of confidence-matched high- and low-KSS views that remain correct under all of the first $k$ random image perturbations, averaged over all 15 MLLMs. Error bars denote the standard deviation across models, and dots show individual models.}
\vspace{-0.3cm}
\label{fig:fig_kss_seeds_bar}
\end{figure}

\begin{figure}[t]
\centering
\includegraphics[width=\linewidth]{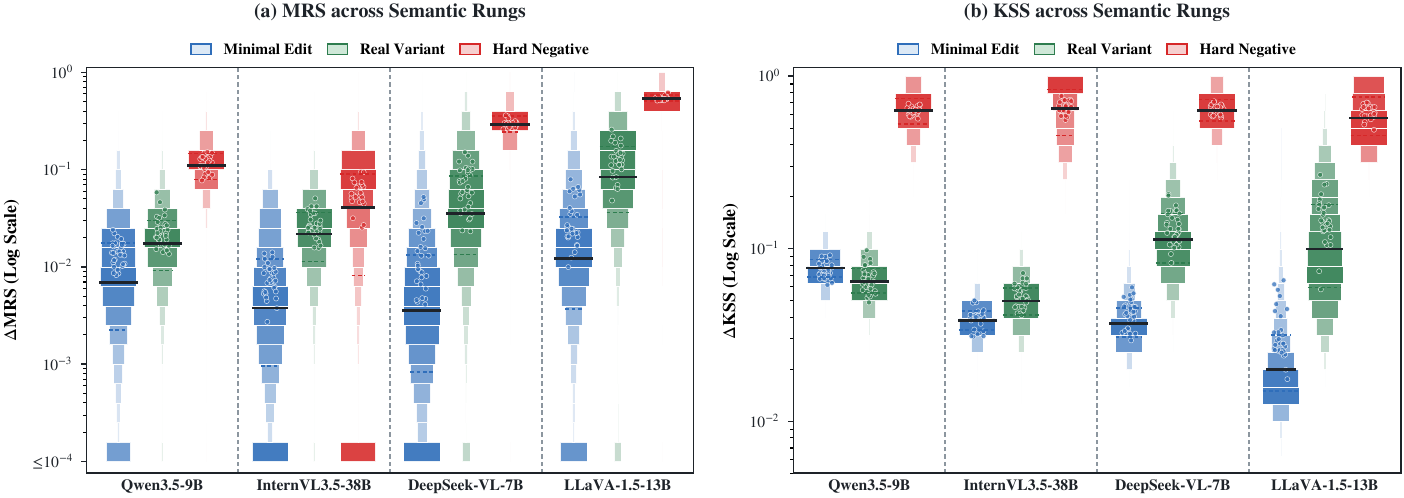}
\vspace{-0.2cm}
\caption{\textbf{Semantic Sensitivity of MRS and KSS.} Changes in MRS and KSS under minimal edits, knowledge-preserving variants, and knowledge-mismatched hard negatives.}
\vspace{-0.3cm}
\label{fig:fig5}
\end{figure}

\noindent\textbf{Qualitative Analysis under Linguistic Variations.}
To examine the influence of multilingual question
formulations~\cite{wang2026m4u}, we analyze translations
of reference questions while keeping the images unchanged.
As shown in the lower half of Figure~\ref{fig:qualitative},
the MRS/PCA panels show that LLaVA-1.5-13B~\cite{liu2024improved} changes its prediction
from found to native under French translation, while Qwen3.5-4B~\cite{qwen3}
changes its prediction from Tribulus to Calliandra under German
translation. The MDS panels accompanying KSS provide two other
examples: LLaVA-1.5-13B~\cite{liu2024improved} changes its prediction from India to
Turkey under Spanish translation, and Qwen3.5-4B~\cite{qwen3} from cultivated
to native under Chinese translation. Together, these cases
illustrate that translating a semantically equivalent question
can alter predictions and their associated internal
representations.

\noindent\textbf{Qualitative Analysis of Correctness-Preserving Variations.}
To avoid equating output correctness with stable latent knowledge, we analyze a knowledge item that remains correctly answered under different perturbations. As illustrated in Figure~\ref{fig:fig_kss_strips}, LLaVA-1.5-13B~\cite{liu2024improved} correctly identifies the museum as located in Brisbane under both a paraphrased question and a closer image viewpoint, with near-maximal MRS in both cases (0.99 and 1.00). However, the two conditions differ in KSS. Under the paraphrase, the perturbed input largely preserves the reference-gradient support profile (KSS 0.97), whereas under the closer viewpoint, a substantial portion of the support falls outside the shared region (KSS 0.43). Thus, although the model produces the same correct answer with comparable output robustness, it exhibits different internal support profiles. Across all models, 12.4\% of correctly answered perturbed inputs in the top MRS quartile fall in the bottom KSS quartile, indicating that such cases are not isolated. These observations suggest that output correctness, and even output-level robustness, cannot fully characterize latent knowledge stability, highlighting the need for evaluation beyond output correctness.

\begin{figure}[t]
\centering
\includegraphics[width=\linewidth]{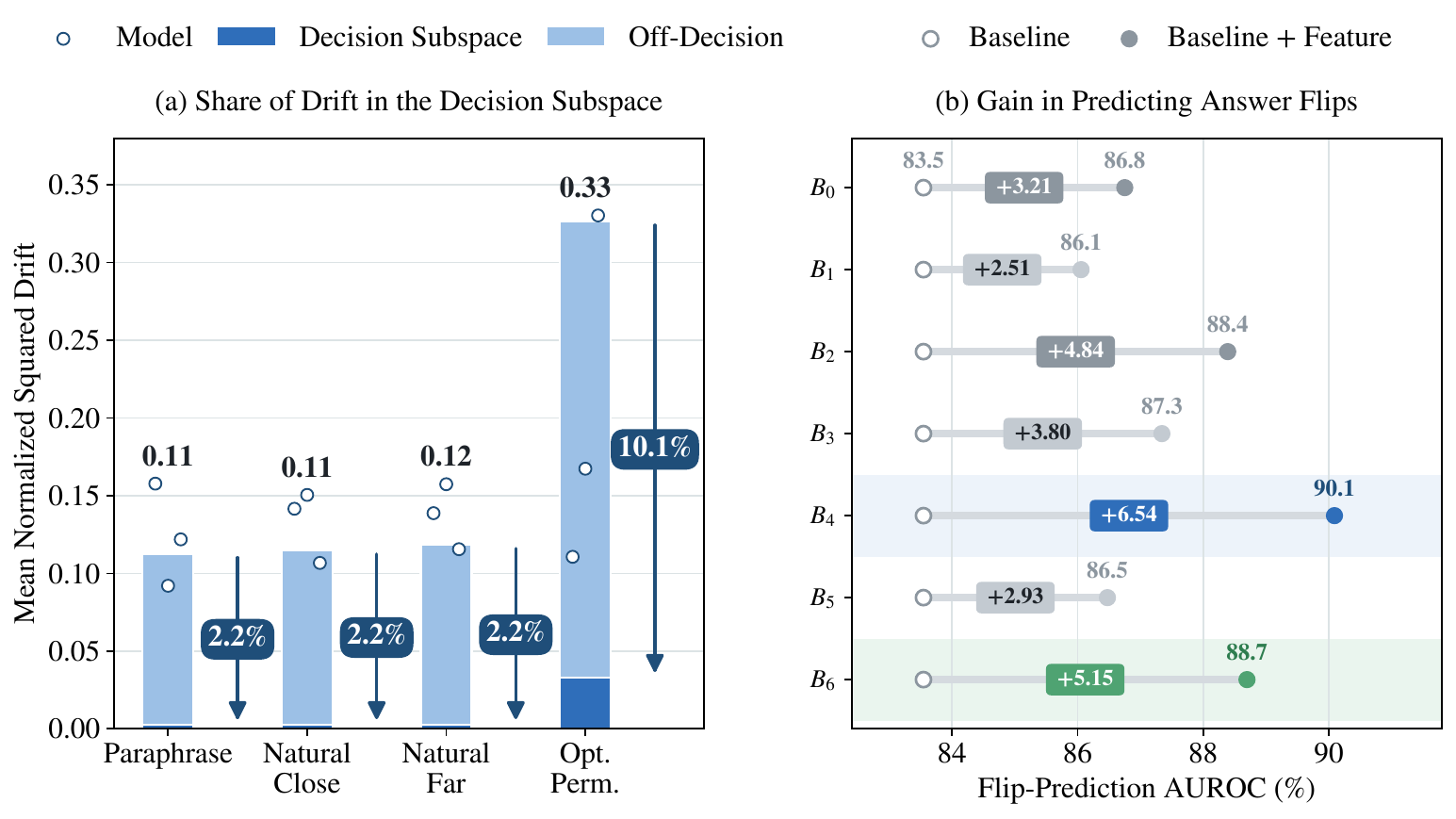}
\vspace{-0.3cm}
\caption{\textbf{Decision-Relevant Representational Drift.} (a) Drift within and outside the decision subspace. (b) Same-condition answer-flip AUROC of a logistic baseline (hollow) and of the baseline augmented with each representation-based feature (filled).}
\vspace{-0.3cm}
\label{fig:fig7}
\end{figure}

\subsection{Validation of MRS and KSS}

\noindent\textbf{Stability and Cross-Condition Correctness.}
To investigate the relationship between latent knowledge stability and output correctness, we divide reference-correct items into high- and low-stability quartiles based on MRS and KSS.
As shown in Figure~\ref{fig:fig4}, higher stability is generally associated with greater accuracy across visual and linguistic elicitation conditions. Specifically, DeepSeek-VL-7B~\cite{lu2024deepseek} exhibits overall accuracy gaps of 18.0\% and 13.5\% between the high- and low-stability groups for MRS and KSS, respectively. Similar trends are observed across other model families, including under different image viewpoints. These findings provide evidence consistent with the correctness–stability hypothesis, suggesting that stable internal knowledge support is associated with more reliable predictions across multimodal elicitation conditions.

\begin{table}[t]
\centering
\caption{\textbf{Effect of Normalization on MRS and KSS.} Spearman's rank correlation coefficients at the model, item, and relation levels under different normalization strategies, computed across 15 models and five target conditions.}
\label{tab:norm_mrs_kss}
\vspace{-0.3cm}
\resizebox{\linewidth}{!}{%
\begin{tabular}{lcccccc}
\toprule
& \multicolumn{3}{c}{\textit{MRS} $\uparrow$}
& \multicolumn{3}{c}{\textit{KSS} $\uparrow$} \\
\cmidrule(lr){2-4} \cmidrule(lr){5-7}
\textbf{Setting}
& Model & Item & Rel.
& Model & Item & Rel. \\
\midrule
No Norm. / Centering
& 0.70 & \textbf{0.92} & \textbf{0.36}
& $-$0.42 & 0.92 & 0.37 \\
Global Norm. / Global Structure
& \textbf{0.88} & \textbf{0.92} & 0.23
& \textbf{0.99} & \textbf{0.99} & \textbf{0.95} \\
\bottomrule
\end{tabular}%
}
\vspace{-0.2cm}
\end{table}

\begin{table}[t]
\centering
\caption{\textbf{Effect of Layer Depth on MRS and KSS.} Spearman's rank correlation coefficients at the model, item, and relation levels across decoder depths and raw hidden states, computed across 15 models and five target conditions.}
\label{tab:depth_mrs_kss}
\vspace{-0.3cm}
\resizebox{\linewidth}{!}{%
\begin{tabular}{lcccccc}
\toprule
& \multicolumn{3}{c}{\textit{MRS} $\uparrow$}
& \multicolumn{3}{c}{\textit{KSS} $\uparrow$} \\
\cmidrule(lr){2-4} \cmidrule(lr){5-7}
\textbf{Setting}
& Model & Item & Rel.
& Model & Item & Rel. \\
\midrule
Depth 50\% Only
& 0.76 & 0.50 & 0.70
& 0.82 & 0.72 & 0.61 \\
Depth 65\% Only
& \textbf{0.90} & 0.75 & 0.87
& \textbf{0.96} & 0.87 & 0.77 \\
Depth 80\% Only
& \textbf{0.90} & \textbf{0.94} & \textbf{0.92}
& 0.81 & \textbf{0.92} & \textbf{0.85} \\
Raw States
& 0.44 & 0.41 & 0.15
& 0.43 & 0.88 & 0.60 \\
\bottomrule
\end{tabular}%
}
\vspace{-0.2cm}
\end{table}

\begin{table}[t]
\centering
\caption{\textbf{Effect of View Removal and Reference Selection.} Spearman's rank correlation coefficients at the model, item, and relation levels under different evaluation settings, relative to the full five-view method (including Translation). Best results are highlighted in \textbf{bold}.}
\label{tab:view_reference_mrs_kss}
\vspace{-0.2cm}
\resizebox{\linewidth}{!}{%
\begin{tabular}{lcccccc}
\toprule
& \multicolumn{3}{c}{\textit{MRS} $\uparrow$}
& \multicolumn{3}{c}{\textit{KSS} $\uparrow$} \\
\cmidrule(lr){2-4} \cmidrule(lr){5-7}
\textbf{Setting}
& Model & Item & Rel.
& Model & Item & Rel. \\
\midrule
\multicolumn{7}{l}{\textcolor{gray}{\textit{Leave-one-view-out}}} \\
Drop Paraphrase
& \textbf{1.00} & 0.94 & 0.88
& \textbf{1.00} & \textbf{0.95} & 0.80 \\
Drop Natural Close
& 0.99 & \textbf{0.96} & \textbf{0.96}
& 0.99 & \textbf{0.95} & \textbf{0.92} \\
Drop Natural Far
& 0.99 & 0.95 & \textbf{0.96}
& 0.99 & \textbf{0.95} & \textbf{0.92} \\
Drop Opt. Perm.
& 0.94 & 0.93 & 0.93
& 0.96 & 0.92 & 0.87 \\
Drop Translation
& 0.91 & 0.87 & 0.90
& 0.92 & 0.89 & 0.87 \\
\midrule
\multicolumn{7}{l}{\textcolor{gray}{\textit{Alternative Reference}}} \\
Paraphrase Reference
& 0.99 & 0.70 & 0.77
& 0.99 & 0.88 & 0.82 \\
Natural Close Reference
& 0.99 & 0.66 & 0.78
& 0.99 & 0.90 & 0.81 \\
\bottomrule
\end{tabular}%
}
\vspace{-0.3cm}
\end{table}

\noindent\textbf{Stability under Repeated Perturbations.}
To investigate whether latent knowledge stability reflects robustness to input noise, we design a repeated-perturbation experiment, in which confidence-matched high- and low-KSS items are required to remain correct across an increasing number of random perturbations. As shown in Figure~\ref{fig:fig_kss_seeds_bar}, high-KSS items remain stable (99.1\% to 97.7\%), whereas low-KSS items degrade steadily (96.9\% to 89.3\%), suggesting that low-stability knowledge is more fragile even when output confidence is comparable.

\noindent\textbf{Semantic Sensitivity of MRS and KSS.}
To investigate whether MRS and KSS distinguish changes in underlying knowledge from surface-level variations, we compare three perturbation types: minimal edits, knowledge-preserving real variants, and knowledge-mismatched hard negatives. As shown in Figure~\ref{fig:fig5}, hard negatives consistently induce substantially larger changes in both MRS and KSS across four representative MLLMs, while minimal edits and real variants generally produce smaller changes. Specifically, this distinction is particularly pronounced for KSS in DeepSeek-VL-7B~\cite{lu2024deepseek} and LLaVA-1.5-13B~\cite{liu2024improved}, indicating greater changes in internal knowledge support when the underlying knowledge differs. These findings suggest that MRS and KSS are sensitive to semantic changes rather than merely surface-level variations, supporting their effectiveness in evaluating latent knowledge stability across multimodal elicitation conditions.

\noindent\textbf{Decision-Relevant Representational Drift.}
To further examine whether MRS and KSS capture decision-relevant rather than generic representational changes, we analyze cross-condition drift and same-condition answer flips. Figure~\ref{fig:fig7}(a) shows that only 2.2\% of drift lies in the decision subspace under paraphrasing and natural-view variations, compared with 10.1\% under option permutations. In Figure~\ref{fig:fig7}(b), each candidate feature is added to the same logistic baseline (hollow markers), which uses the log reference margin and model, condition, and question-template indicators (83.5\% AUROC). $B_0$ denotes raw representational drift; $B_1$ denotes random four-dimensional subspace drift; and $B_2$ denotes decision-subspace drift. $B_3$ is an MRS control using the Jacobian from a mismatched item, while $B_4$ is the proposed MRS. Likewise, $B_5$ is a KSS control using a mismatched-item margin gradient, while $B_6$ is the proposed KSS. Restricting drift to the decision subspace ($B_2$, 88.4\%) is more informative than drift in a random subspace of the same dimension ($B_1$, 86.1\%). MRS ($B_4$) achieves 90.1\% AUROC, compared with 87.3\% for $B_3$, while KSS ($B_6$) achieves 88.7\%, compared with 86.5\% for $B_5$. Both also outperform raw drift ($B_0$, 86.8\%), further supporting the importance of decision-grounded stability measurements beyond generic overall representation changes. See the appendix for more results.

\begin{figure}[t]
\centering
\includegraphics[width=\linewidth]{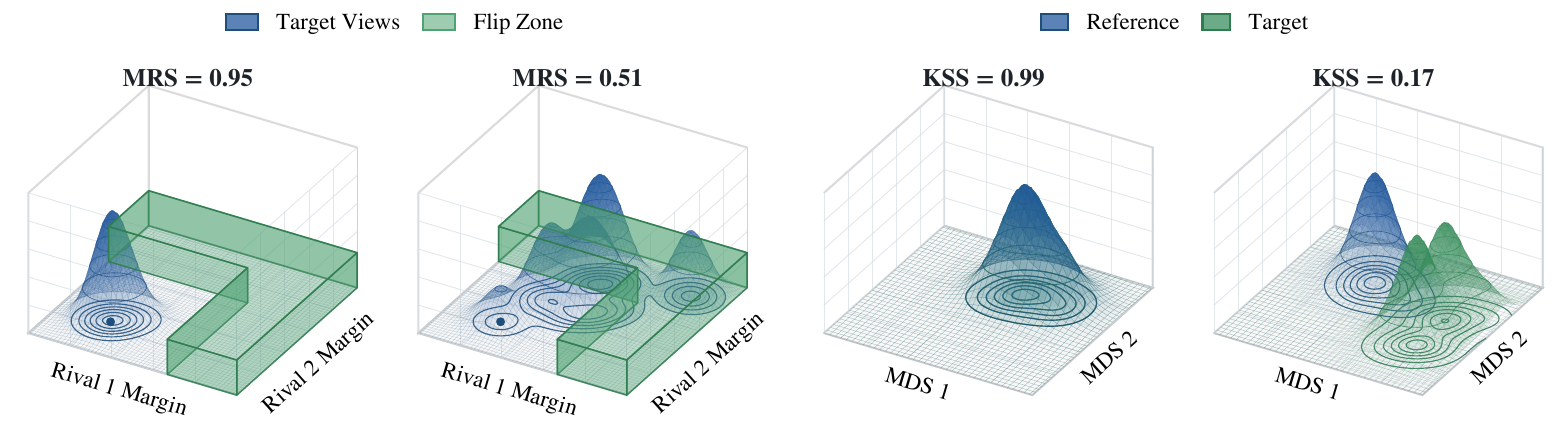}
\vspace{-0.3cm}
\caption{\textbf{Geometric Interpretation of MRS and KSS.} High- and low-stability examples illustrate differences in prediction margins and internal knowledge support distributions.}
\vspace{-0.5cm}
\label{fig:fig8}
\end{figure}

\begin{figure}[t]
\centering
\includegraphics[width=\linewidth]{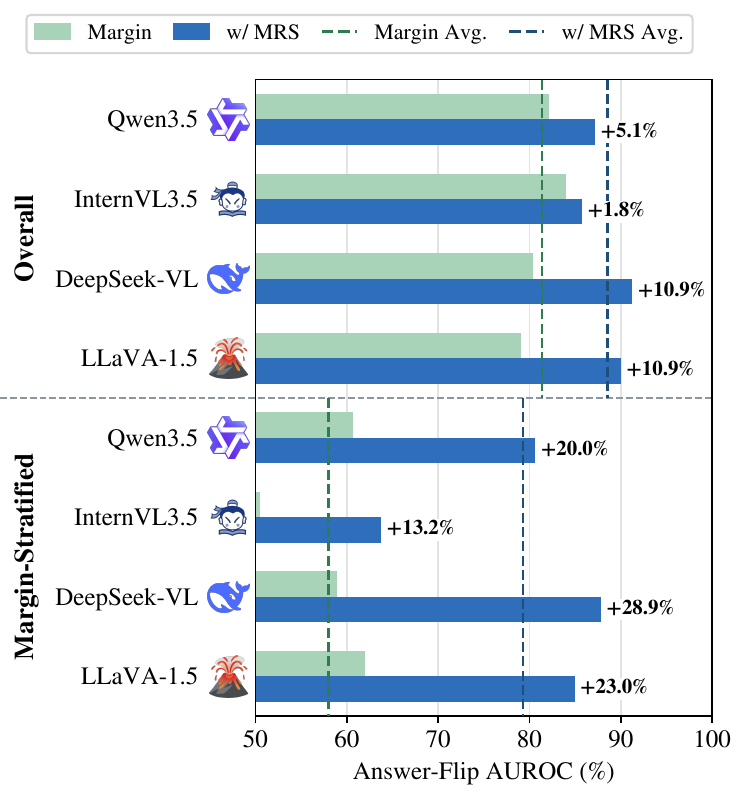}
\vspace{-0.3cm}
\caption{\textbf{Flip Characterization Beyond Confidence.} Answer-flip AUROC with and without MRS under overall and margin-stratified evaluations.}
\vspace{-0.5cm}
\label{fig:fig9}
\end{figure}

\noindent\textbf{Effect of Normalization.}
To investigate the influence of normalization on latent knowledge stability evaluation, we compare different normalization strategies for MRS and KSS. As shown in Table~\ref{tab:norm_mrs_kss}, global normalization improves model-level results for MRS from 0.70 to 0.88 and KSS from $-$0.42 to 0.99, while relation-level MRS decreases from 0.36 to 0.23. These results suggest that normalization affects the two metrics differently across evaluation levels.

\noindent\textbf{Effect of Layer Depth.}
To examine how decoder depth affects stability evaluation, we compare MRS and KSS across layers and raw hidden states. As shown in Table~\ref{tab:depth_mrs_kss}, deeper layers generally yield higher values, with item-level MRS increasing from 0.50 to 0.94 and KSS from 0.72 to 0.92 between 50\% and 80\% decoder depth. These findings suggest that later decoder layers~\cite{tang2026diagnosing} provide more informative representations for evaluating latent knowledge stability.

\noindent\textbf{Effect of Elicitation Conditions and Reference Selection.}
To assess the robustness of MRS and KSS, we evaluate their sensitivity to removing individual elicitation conditions and changing reference inputs. As shown in Table~\ref{tab:view_reference_mrs_kss}, both metrics maintain high model-level results under most settings, while removing option permutations reduces model-level MRS to 0.94. Alternative reference conditions also produce larger variations in item-level MRS. These findings highlight the relative robustness of aggregate stability measurements while revealing sensitivity to reference selection and specific elicitation conditions.

\subsection{Deeper Analysis}

\noindent\textbf{Geometric Interpretation of MRS and KSS.}
To intuitively illustrate the complementary aspects of latent knowledge stability captured by MRS and KSS, we visualize representative high- and low-stability examples. As shown in Figure~\ref{fig:fig8}, high-MRS samples (0.95) remain concentrated near the reference prediction and away from the answer-flip region, whereas low-MRS samples (0.51) exhibit larger shifts toward or across the decision boundary. Similarly, high-KSS samples (0.99) preserve nearly identical internal support distributions across elicitation conditions, while low-KSS samples (0.17) show substantial redistribution of internal knowledge support. These observations demonstrate that MRS captures the robustness of prediction preferences, whereas KSS characterizes the consistency of internal knowledge support, highlighting two complementary dimensions of latent knowledge stability in MLLMs.

\noindent\textbf{Flip Characterization Beyond Confidence.}
To investigate whether MRS provides additional information beyond the model's reference prediction confidence, we evaluate its ability to discriminate answer flips across four MLLMs. As shown in Figure~\ref{fig:fig9}, incorporating MRS improves the average answer-flip AUROC from 81.4\% to 88.6\%, with the largest gains observed in DeepSeek-VL~\cite{lu2024deepseek} and LLaVA-1.5~\cite{liu2024improved} (+10.9\% each). Furthermore, when controlling for the reference prediction margin through stratification, MRS achieves an average AUROC of 79.3\%, compared with only 58.0\% for the margin alone, with DeepSeek-VL~\cite{lu2024deepseek} showing the largest improvement (+28.9\%). These findings demonstrate that MRS captures cross-condition prediction robustness beyond the reference prediction margin, providing additional information about the sensitivity of MLLM predictions to multimodal elicitation conditions.

\begin{figure}[t]
\centering
\includegraphics[width=\linewidth]{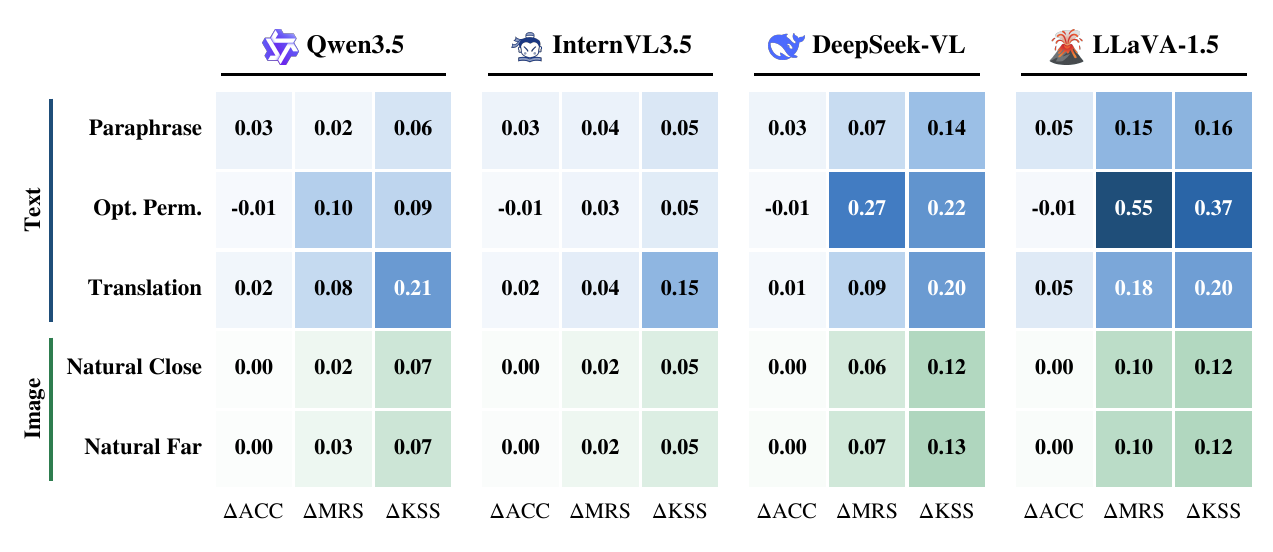}
\vspace{-0.3cm}
\caption{\textbf{Sensitivity to Elicitation Conditions.}
$\Delta$ACC is reference minus target accuracy over all 19{,}452 items; $\Delta$MRS and $\Delta$KSS average $1-\mathrm{score}$ over reference-correct items.}
\vspace{-0.5cm}
\label{fig:fig10}
\end{figure}

\noindent\textbf{Sensitivity to Elicitation Conditions.}
To investigate how elicitation conditions affect latent knowledge stability across model families, we compare changes in ACC, MRS, and KSS under visual and linguistic variations. As shown in Figure~\ref{fig:fig10}, ACC remains largely unchanged across conditions, whereas MRS and KSS reveal substantial model-dependent differences. Under option permutations, LLaVA-1.5~\cite{liu2024improved} exhibits decreases of 0.55 in MRS and 0.37 in KSS, compared with 0.27 and 0.22 for DeepSeek-VL~\cite{lu2024deepseek}. In contrast, visual variations generally induce smaller changes in MRS but more noticeable changes in KSS. These findings reveal model-dependent sensitivities of MRS and KSS to visual and linguistic elicitation conditions.

\section{Application}

\begin{table}[t]
\centering
\caption{\textbf{Effect of an MRS-Inspired Regularizer on Multimodal Knowledge Editing.} Token-level accuracy (\%) on visual entity editing instances of MMKE-Bench using LLaVA-1.5-7B, reporting text reliability (T-Rel), image reliability (I-Rel), image generalization (I-Gen), text locality (T-Loc), image locality (I-Loc), and portability (Port).}
\label{tab:visual_entity_editing}
\resizebox{\linewidth}{!}{%
\begin{tabular}{lcccccc}
\toprule
Method & T-Rel $\uparrow$ & I-Rel $\uparrow$ & I-Gen $\uparrow$ & T-Loc $\uparrow$ & I-Loc $\uparrow$ & Port $\uparrow$ \\
\midrule
FT-LLM~\cite{cheng2023can}
& 53.89 & 49.54 & 49.30 & 77.71 & 17.58 & 41.23 \\
\rowcolor{altrow}\textbf{+ Ours}
& \textbf{55.13} & \textbf{52.98} & \textbf{53.06} & \textbf{77.85} & \textbf{18.62} & \textbf{41.71} \\
\midrule
KE~\cite{cheng2023can}
& 46.45 & 46.19 & 46.29 & 79.89 & 18.73 & 48.77 \\
\rowcolor{altrow}\textbf{+ Ours}
& \textbf{48.51} & \textbf{48.38} & \textbf{48.33} & \textbf{81.31} & \textbf{19.72} & \textbf{48.96} \\
\bottomrule
\end{tabular}%
}
\vspace{-0.3cm}
\end{table}

\noindent\textbf{Setting.}
To investigate whether MRS can improve knowledge editing generalization in MLLMs, we incorporate an MRS-inspired margin-retention regularizer into existing knowledge editing methods. Conventional approaches often succeed on the original editing image but fail to preserve the edited knowledge~\cite{li2026multimodal} across different images of the same entity. To address this limitation, we generate multiple knowledge-preserving views from the original editing image and encourage the model to maintain the reference answer margin across these views. Specifically, we integrate this regularizer into FT-LLM~\cite{cheng2023can} and KE~\cite{cheng2023can} without requiring additional training data or retraining the editing hypernetwork. We evaluate both methods on 955 visual entity editing instances~\cite{jiang2025mmke} using LLaVA-1.5~\cite{liu2024improved}, reporting six metrics: text reliability (T-Rel), image reliability (I-Rel), image generalization (I-Gen), text locality (T-Loc), image locality (I-Loc), and portability (Port).

\noindent\textbf{Results.}
From Table~\ref{tab:visual_entity_editing}, the MRS-inspired regularizer improves reliability and generalization for both editing methods. For FT-LLM, T-Rel, I-Rel, and I-Gen increase by 1.24, 3.44, and 3.76 points, respectively, with larger gains in image-related metrics. For KE, the regularizer yields balanced improvements of 2.06, 2.19, and 2.04 points across the three metrics. These gains are accompanied by consistent improvements in both text and image locality (by 0.14--1.42 points), while portability remains largely unchanged. These results suggest that the regularizer enhances editing reliability and generalization across different images of the same entity, with consistent benefits across editing methods. More details are provided in the Appendix.

\section{Conclusion}

In this paper, we systematically investigate latent knowledge stability in MLLMs beyond output correctness. We introduce two complementary metrics, margin retention score (MRS) and latent knowledge support-structure stability (KSS), to evaluate prediction robustness and the consistency of internal knowledge support across visual and linguistic elicitation conditions. We further construct MMStableBench, comprising 19,452 knowledge items, to enable systematic evaluation across different MLLMs. Extensive experiments reveal that output correctness and model scaling do not necessarily reflect latent knowledge stability, while higher internal stability is generally associated with more reliable predictions across elicitation conditions. Furthermore, incorporating an MRS-inspired regularizer into existing knowledge editing methods improves the generalization of edited knowledge across different images and question formulations. Our findings highlight the importance of evaluating latent knowledge stability for a more comprehensive understanding of knowledge representation and utilization in MLLMs.

\bibliographystyle{plainnat}
\bibliography{main}

\clearpage

\onecolumn
\beginappendix


The appendix is organized as follows. Appendix~\ref{app:notation} summarizes the main notations used in our work, while Appendix~\ref{more_ex} provides detailed experimental protocols and implementation settings. Appendix~\ref{exp} presents additional qualitative analyses and geometric visualizations of MRS and KSS across different MLLMs. Appendix~\ref{lim} discusses the limitations and potential future extensions of our work. Finally, Appendix~\ref{dis} provides further discussions on the design and interpretation of the proposed metrics.

\section{Notation Summary}
\label{app:notation}

For clarity, Table~\ref{tab:notation-summary} summarizes the main notation used in our work, covering the problem setup, MRS, KSS, semantic sensitivity analysis, and the knowledge editing application.

\begin{table}[h]
\centering
\caption{\textbf{Summary of the main notation used in our work.}}
\label{tab:notation-summary}
\footnotesize
\renewcommand{\arraystretch}{0.90}
\begin{tabular}{p{0.26\linewidth}p{0.53\linewidth}}
\toprule
\textbf{Notation} & \textbf{Description} \\
\midrule
$i,\,j$ & Knowledge-item indices ($j\neq i$ for a mismatched item) \\
$m$ & MLLM index \\
$K_i$ & Number of answer options for knowledge item $i$ \\
$a,\,t,\,c$ & Reference, target, and generic elicitation-condition indices \\
$\mathcal T_i$ & Set of alternative elicitation conditions for item $i$ \\
$x_{i,a},\,x_{i,t}$ & Reference and target inputs for item $i$ \\
$y_i$ & Verified answer to knowledge item $i$ \\
$\hat y_{m,i}$ & Model $m$'s prediction under the reference condition \\
$\boldsymbol z_{m,i,c}$ & Answer-choice logit vector under condition $c$ \\
$\boldsymbol h_{m,i,c}^{\ell}$ & Last-prompt-token residual state at decoder layer $\ell$ \\
$\mathcal L_m$ & Set of evaluated decoder layers for model $m$ \\
\midrule
$k$ & Index of an answer option \\
$g_{m,i,k}$ & Floored reference logit margin over competing option $k$ \\
$\tau$ & Positive margin floor (set to $0.5$) \\
$\Delta\boldsymbol h_{m,i,t}^{\ell}$ & Target-minus-reference residual-state displacement at layer $\ell$ \\
$J_{m,i,a}^{\ell}$ & Reference Jacobian of answer-choice logits with respect to the layer-$\ell$ state \\
$\boldsymbol\delta_{m,i,t}^{\ell}$ & First-order logit change induced by the residual-state displacement \\
$\mathcal D_{m,i}(\boldsymbol\delta)$ & Normalized, one-sided degradation of the reference prediction margin \\
$\operatorname{MRS}_{m,i,t}$ & Margin retention score for item $i$ under target condition $t$ \\
\midrule
$r_{m,i}$ & Strongest rival to the verified answer under the reference condition \\
$\boldsymbol\gamma_{m,i}^{\ell}$ & Fixed reference gradient of the verified-answer margin at layer $\ell$ \\
$D_{\ell}$ & Hidden-state width at decoder layer $\ell$ \\
$q,\,b$ & Hidden-dimension and channel-block indices \\
$\mathcal B_{\ell,b}$ & Set of hidden dimensions in channel block $b$ at layer $\ell$ \\
$d_b^{\ell}(x)$ & Nonnegative reference-gradient projection of channel block $b$ for input $x$ \\
$M^{\ell}(c)$ & Total positive support mass at layer $\ell$ under condition $c$ \\
$\epsilon$ & Support detection threshold (set to $10^{-12}$) \\
$\mathcal L^{\mathrm{id}}$ & Set of layers with identifiable support in both conditions \\
$\operatorname{KSS}^{\ell}$ & Weighted Dice overlap of reference and target support profiles at layer $\ell$ \\
$\operatorname{KSS}$ & Mean support-structure stability over identifiable layers \\
\midrule
$y^{\ast}$ & Edited target answer that the model should produce \\
$x,\,\{x_v\}_{v=1}^{V}$ & Editing input and its $V$ knowledge-preserving views \\
$m(u)$ & Average target-token logit margin for input $u$ \\
$\mathrm{sg}(\cdot)$ & Stop-gradient operator \\
$\mathcal L_{\mathrm{MR}}$ & MRS-inspired margin-retention regularizer \\
$\mathcal L_{\mathrm{NLL}}$ & Negative log-likelihood loss for the edited target \\
$\lambda$ & Weight of the margin-retention regularizer \\
$\mathcal L$ & Total knowledge-editing objective \\
\bottomrule
\end{tabular}
\end{table}

\section{More Experiment Details}
\label{more_ex}

\subsection{More Details about Protocol}
Scores are first averaged across variants of each knowledge item and then across items. The model-family and scaling comparison in the main text reports benchmark-level stability, whereas the correctness–stability hypothesis focuses specifically on reference-correct items. Accordingly, all hypothesis tests are restricted to items answered correctly under the reference condition, for both MRS and KSS.

For the stability and cross-condition correctness experiment in the main text, we adopt a leave-one-condition-out protocol over five elicitation conditions: paraphrase, natural close, natural far, option permutation, and translation. Each condition is held out in turn, and an item's stability is computed as the mean MRS or KSS over the remaining four conditions. For each model, metric, and held-out condition, items at or below the 25th percentile and at or above the 75th percentile are assigned to the low- and high-stability groups, respectively. Their accuracies are then compared on the held-out condition. This experiment evaluates cross-condition correctness using held-out conditions, whereas the decision-relevant representational drift and flip characterization experiments in the main text analyze answer flips on the same condition used to compute MRS.

For the ablation experiments on normalization, layer depth, elicitation-condition removal, and reference selection in the main text, each reported value is the Spearman correlation between a metric variant and the corresponding full metric. Model-level correlations are computed across the mean scores of 15 models. Item-level correlations are computed over item-condition scores within each model and then averaged across models. Relation-level correlations are computed over question-template means within each model, using templates containing at least 30 items as relation proxies, and then averaged across models. Higher correlations indicate closer agreement with the full metric. Confidence intervals are estimated using bootstrap resampling at the knowledge-item level.

\subsection{Implementation Details of KSS}
\label{app:kss-details}

\noindent\textbf{Layers and states.} For each model, MRS and KSS use the hidden states of the last prompt token~\cite{zhang2025cross} at three decoder layers of the language model located at 50\%, 65\%, and 80\% of the network depth. For a model with $L$ decoder layers, the zero-based layer index for depth fraction $r\in\{0.5,0.65,0.8\}$ is $\lfloor rL\rfloor-1$. This yields layers $\{11,14,18\}$ for the 24-layer models (Qwen3.5-0.8B~\cite{qwen3}, Qwen3.5-2B~\cite{qwen3}, and DeepSeek-VL-1.3B~\cite{lu2024deepseek}), $\{13,17,21\}$ for the 28-layer models (InternVL3.5-1B~\cite{wang2025internvl3} and 2B), $\{14,18,23\}$ for DeepSeek-VL-7B~\cite{lu2024deepseek} ($L=30$), $\{15,19,24\}$ for the 32-layer models (Qwen3.5-4B~\cite{qwen3}, Qwen3.5-9B~\cite{qwen3}, and LLaVA-1.5-7B~\cite{liu2024improved}), $\{17,22,27\}$ for the 36-layer models (InternVL3.5-4B~\cite{wang2025internvl3} and 8B), $\{19,25,31\}$ for the 40-layer models (InternVL3.5-14B~\cite{wang2025internvl3} and LLaVA-1.5-13B~\cite{liu2024improved}), and $\{31,40,50\}$ for the 64-layer models (Qwen3.5-27B~\cite{qwen3} and InternVL3.5-38B~\cite{wang2025internvl3}). The reference gradient $\boldsymbol\gamma^{\ell}_{m,i}$ is the gradient of the restricted-logit margin between the verified answer and its strongest rival, computed once under the reference condition and held fixed for all conditions of the same knowledge item.

\noindent\textbf{Channel blocks.} At each selected layer, the $D_\ell$ hidden dimensions are partitioned in their original order into 64 contiguous, non-overlapping, equal-width channel blocks, so each block contains $D_\ell/64$ dimensions: 16 for $D_\ell=1024$ (Qwen3.5-0.8B~\cite{qwen3} and InternVL3.5-1B~\cite{wang2025internvl3}), 32 for $D_\ell=2048$ (Qwen3.5-2B~\cite{qwen3}, InternVL3.5-2B~\cite{wang2025internvl3}, and DeepSeek-VL-1.3B~\cite{lu2024deepseek}), 40 for $D_\ell=2560$ (Qwen3.5-4B~\cite{qwen3} and InternVL3.5-4B~\cite{wang2025internvl3}), 64 for $D_\ell=4096$ (Qwen3.5-9B~\cite{qwen3}, InternVL3.5-8B~\cite{wang2025internvl3}, DeepSeek-VL-7B~\cite{lu2024deepseek}, and LLaVA-1.5-7B~\cite{liu2024improved}), and 80 for $D_\ell=5120$ (Qwen3.5-27B, InternVL3.5-14B, InternVL3.5-38B~\cite{wang2025internvl3}, and LLaVA-1.5-13B~\cite{liu2024improved}). The contributions $\gamma^{\ell}_q h^{\ell}_q(x)$ are summed within each block before applying the positive part to obtain $d^{\ell}_b(x)$, which approximates the first-order effect~\cite{kramar2024atp} of jointly removing block $b$ on the verified-answer margin.

\noindent\textbf{Detection threshold and indeterminate items.} We set $\epsilon=10^{-12}$, which only excludes layers whose positive support is numerically zero. If no layer is identifiable for a reference--target pair ($|\mathcal L^{\mathrm{id}}|=0$), the layer-averaged KSS is undefined; such a pair is treated as indeterminate, assigned no KSS value, and excluded from all KSS statistics rather than imputed.

\noindent\textbf{Coverage.} Across all 15 models and all elicitation conditions, this case never occurs. Among the 231{,}132 items answered correctly under the reference condition, all 1{,}155{,}660 reference--target pairs have at least one identifiable layer, so KSS coverage is 100\% for every model and condition. Only 61 pairs (0.005\%) have fewer than three identifiable layers, and their KSS is averaged over the remaining identifiable layers. No pair is excluded from KSS statistics, although individual layers may be excluded. In 59 of these pairs (14 for DeepSeek-VL-7B~\cite{lu2024deepseek} and 45 for LLaVA-1.5-13B~\cite{liu2024improved}), a layer is excluded because support vanishes in only one profile. As a sensitivity check, assigning such layers a score of 0 (complete support loss) instead of excluding them changes any model-level KSS by at most $2.1\times10^{-4}$ and alters only one entry of Table~1 in the main paper at the third decimal (DeepSeek-VL-7B~\cite{lu2024deepseek}, Option Perm.: 0.778 $\rightarrow$ 0.777).

\subsection{Construction of the Semantic Sensitivity Analysis}
\label{app:semantic-sensitivity}

\noindent\textbf{Perturbation types.}
The semantic sensitivity analysis uses only items answered correctly under the reference condition, and evaluates three types of changes with increasing semantic distance from the reference input. Minimal edits leave the question, options, and image content unchanged: we either prepend the neutral instruction ``Please look at the image carefully.'' to the prompt or re-encode the reference image as JPEG with quality 90, and average the two per item. Real variants are knowledge-preserving changes, namely question paraphrasing and the natural close and far views of the same entity, averaged per item. Option permutation is excluded here to isolate pure semantic variations from sensitivity to answer-label reassignment (discussed in Appendix~\ref{dis}). Hard negatives pair each item $i$ with a donor item $j\neq i$ whose question begins with the same three words, used as a proxy for the queried relation (e.g., ``what family does''), and whose entity differs from that of item $i$. The donor is selected deterministically by a hash-based offset within this pool, and the hard-negative input is the donor's own reference input $x_{j,a}$, with its own image, question, and four answer options. To prevent overlap in answer labels from diluting the comparison, we retain only pairs in which the correct options of the two items carry different letters, yielding 11{,}493 of the 19{,}452 items. In the retained pairs, the entities and images always differ, the questions are identical after masking the entity noun in 12.9\% of pairs, and the correct answer texts coincide in 11.5\% of pairs.

\noindent\textbf{Metric definitions.}
Since an exact repeat of the reference input yields a score of one under both metrics, we define $\Delta\mathrm{MRS}=1-\mathrm{MRS}$ and $\Delta\mathrm{KSS}=1-\mathrm{KSS}$. For hard negatives, MRS projects the displacement between the donor's and item $i$'s reference hidden states through item $i$'s fixed reference Jacobian, and KSS compares the reference support of item $i$ with that of donor $j$, both computed with item $i$'s fixed gradient $\boldsymbol\gamma^{\ell}_{m,i}$ and channel partition. Both metrics are therefore evaluated with respect to item $i$'s verified answer and strongest rival, and the option spaces of the two items are never mixed. Both panels use a logarithmic axis: $\Delta\mathrm{MRS}$ is exactly zero for some items, mostly under minimal edits, so its values are floored at $10^{-4}$, whereas $\Delta\mathrm{KSS}$ has no zeros. Each distribution is shown as a log-binned histogram whose width reflects density; solid and dashed lines mark the median and quartiles, and dots show question-template means for templates with at least 20 items. Paired one-sided Wilcoxon signed-rank tests within items confirm that hard negatives yield larger changes than real variants for both metrics and all four models ($p<10^{-6}$ at both the item and template levels), and that minimal edits yield smaller changes than real variants in all cases except KSS for Qwen3.5-9B, where the two are comparable.

\noindent\textbf{Scope of the conclusion.}
Hard negatives differ from the reference in image and entity, and in most cases also in the question wording and correct answer, whereas minimal edits and real variants preserve the underlying knowledge. The ordering of $\Delta\mathrm{MRS}$ and $\Delta\mathrm{KSS}$ across the three types is therefore consistent with sensitivity to changes in the underlying knowledge, and we interpret these results as evidence that both metrics are semantically sensitive.

\subsection{Evaluation Protocols for Answer-Flip Analyses}
\label{app:flip-protocols}

\noindent\textbf{Two protocols.} Our answer-flip analyses follow two protocols that support different conclusions. The stability and cross-condition correctness experiment in the main paper uses a leave-one-condition-out protocol: for each item and held-out condition, stability is computed only from the remaining conditions, so the held-out condition never contributes to its own score. This setting evaluates whether stability measured on observed conditions anticipates errors on an unobserved condition. In contrast, the decision-relevant representational drift and flip characterization experiments in the main paper compute MRS on the same condition whose answer flip is evaluated, with KSS additionally included in the former. These analyses therefore assess same-condition flip discrimination: whether the metrics characterize and discriminate the conditions on which the answer changes, given access to the target condition. They should not be interpreted as evidence that errors can be predicted before the target condition is observed.

\noindent\textbf{Decision-relevant representational drift.} For every item answered correctly under the reference condition and each elicitation condition, the label indicates whether the answer under that condition flips. The baseline is a logistic regression with model, condition, and question-template indicators and the log reference margin as features; each measure (raw representational drift, MRS, KSS, and their mismatched-gradient controls) is added to this baseline separately. Predictions are obtained by 5-fold cross-validation with folds grouped by knowledge item, so that all conditions of an item fall in the same fold, and AUROC is computed only on held-out-fold predictions. 95\% confidence intervals are obtained by item-level bootstrap.

\noindent\textbf{Decision subspace and drift features.} For item $i$ and each selected layer $\ell$, the decision subspace is the span of the four rows of the reference Jacobian $J_{m,i,a}^{\ell}$, one per answer option, so its dimension is at most four; an orthonormal basis is obtained by QR decomposition. The random-subspace control uses a Gaussian matrix of the same shape whose rows are rescaled to the norms of the corresponding Jacobian rows, so that it also spans a four-dimensional subspace, with a fixed per-item seed. Squared drift norms are divided by $\beta_\ell^2$, where $\beta_\ell$ is the median pairwise distance between reference hidden states for the given model and layer, and are averaged over the three selected layers. In Figure~8(a), the displacement $\Delta h^{\ell}_{m,i,t}$ is split exactly into its projection onto the decision subspace and the orthogonal complement, and the percentages give the decision-subspace share of the total squared drift. In Figure~8(b), each feature is added separately to the logistic baseline described above, after z-scoring within each model. The Baseline row adds the normalized raw drift norm $\|\Delta h\|$. $B_0$ adds the norm of the raw displacement projected onto the random four-dimensional subspace. $B_1$ adds the norm of the decision-subspace projection of the item-specific residual, obtained by subtracting the condition-level mean displacement estimated on the other folds. $B_2$ and $B_4$ are MRS and KSS computed with the Jacobian and decision gradient of a different same-relation item, respectively, and $B_3$ and $B_5$ are MRS and KSS themselves; all stability scores enter as $1-\text{score}$. The analysis uses Qwen3.5-9B, InternVL3.5-38B, DeepSeek-VL-7B, and LLaVA-1.5-13B, and the reported AUROC corresponds to the baseline augmented with each single feature.

\noindent\textbf{Flip characterization beyond confidence.} For each of the four models, we use 500 randomly sampled knowledge items with four conditions (paraphrase, natural close view, natural far view, and option permutation), and label a condition as flipped if its prediction differs from the reference prediction. In the overall evaluation, the baseline is a logistic regression on the log reference margin and condition indicators, and the full model additionally includes the same-condition MRS; both are evaluated with item-grouped 5-fold cross-validation, and AUROC is computed on held-out-fold predictions. In the margin-stratified evaluation, no model is fitted: items are divided into quintiles of the reference margin, AUROC is computed directly from the MRS ranking within each quintile, and the results are averaged over quintiles, with the margin itself evaluated in the same way as a reference. 95\% confidence intervals are obtained by item-level bootstrap with 2{,}000 resamples.

\subsection{Details of the Knowledge Editing Application}
\label{app:editing}

\noindent\textbf{Edited target.} In knowledge editing, the reference answer is the edited target answer $y^{\ast}$ that the model is required to produce after editing, rather than the model's original prediction before editing. The regularizer therefore encourages the edited knowledge, not the pre-edit answer, to be retained across views.

\noindent\textbf{MRS-inspired margin-retention regularizer.} Let $x$ denote the editing input (editing image and question) and $\{x_v\}_{v=1}^{V}$ its views. For an input $u$, we define the target margin as the average, over the answer tokens of $y^{\ast}$, of the logit of the target token minus the largest logit among all other vocabulary tokens:
\begin{equation}
m(u)=\frac{1}{|y^{\ast}|}\sum_{s=1}^{|y^{\ast}|}\Big(z_s(u)_{y^{\ast}_s}-\max_{k\neq y^{\ast}_s} z_s(u)_k\Big).
\end{equation}
Following the margin-degradation form of MRS, the regularizer penalizes only reductions of the target margin on each view relative to the editing input:
\begin{equation}
\mathcal L_{\mathrm{MR}}=\frac{1}{V}\sum_{v=1}^{V}\frac{\big[\,\mathrm{sg}(m(x))-m(x_v)\,\big]_+}{\max\{\mathrm{sg}(m(x)),\,0.5\}},
\end{equation}
where $\mathrm{sg}(\cdot)$ denotes stop-gradient, so that only the margins on the views are increased toward that of the editing input, and the floor of 0.5 follows the minimum margin used in MRS. Unlike the evaluation metric, the regularizer uses the actual forward-pass margins on the views rather than the first-order approximation. The editing objective is
\begin{equation}
\mathcal L=\mathcal L_{\mathrm{NLL}}(y^{\ast}\mid x)+\lambda\,\mathcal L_{\mathrm{MR}}.
\end{equation}

\noindent\textbf{Integration into editing methods.} For FT-LLM~\cite{cheng2023can}, we fine-tune the up- and down-projection weights of the MLP in the last decoder layer (layer 31) with AdamW for 40 steps at a learning rate of $10^{-4}$, following the default configuration, and optimize $\mathcal L$ instead of $\mathcal L_{\mathrm{NLL}}$. For KE~\cite{cheng2023can}, we use the officially released hypernetwork checkpoint without retraining; the regularizer enters through the gradient consumed by the hypernetwork, which is computed from $\mathcal L$ on the unedited model instead of from $\mathcal L_{\mathrm{NLL}}$ alone.

\noindent\textbf{View generation.} The $V=4$ views are generated only from the editing image of each instance: horizontal flipping, a 80\% center crop, a 85\% top-left crop (both resized to the original resolution), and a $5\times5$ box blur. The question and target answer are unchanged. No evaluation images, rephrased questions, or other externally annotated data are used to construct the views.

\noindent\textbf{Hyperparameter selection.} The weight $\lambda$ is selected from $\{0.3,1,3\}$ for FT-LLM and $\{0.1,0.3,1,3\}$ for KE by maximizing the mean of the three metrics on 60 randomly sampled editing instances from the training split, yielding $\lambda=3$ for FT-LLM and $\lambda=1$ for KE. The 955 evaluation instances are used only once for the final evaluation.

\noindent\textbf{Model and metrics.} All experiments use LLaVA-1.5-7B~\cite{liu2024improved} on the visual entity editing subset of MMKE-Bench~\cite{jiang2025mmke}. FT-LLM fine-tunes parameters of the language model, and KE (KnowledgeEditor) predicts parameter updates with a trained hypernetwork. 
Following MMKE-Bench, we report token-level accuracy
(\%) for six metrics: T-Rel (text reliability),
I-Rel (image reliability), I-Gen (image generalization),
T-Loc (text locality), I-Loc (image locality),
and Port (portability), accuracy on questions about the edited knowledge with the editing image; and I-Gen (image generalization), accuracy on the edited knowledge with a different
image of the same entity.

\section{More Experiments}
\label{exp}

\subsection{Channel Partitioning Sensitivity}
To verify the robustness of latent knowledge support-structure stability (KSS) to hidden-dimension ordering and channel-block granularity, we conduct ablation experiments with random channel permutations (across 10 independent random seeds) and multi-scale block counts ($B \in \{16, 32, 64, 128, 256\}$) across all 15 MLLMs and five target elicitation conditions, totaling 1,155,660 reference–target pairs. The results show that KSS is highly invariant to the coordinate ordering of residual-stream hidden dimensions and maintains strong item-level consistency, model-ranking stability, and cross-condition accuracy predictive power across block granularities spanning two orders of magnitude. Specifically, under 10 random channel permutations ($B=64$), the mean item-level Spearman correlation across the 15 models reaches 0.968 (with a single-model minimum of 0.940, and reaching 0.992 on LLaVA-1.5-7B/13B and 0.989 on DeepSeek-VL-1.3B), while the mean model-level ranking correlation reaches 0.990 (up to 1.000, with a mean pairwise ranking correlation of 0.981 among the 10 seeds and cross-seed standard deviations of model-mean KSS of only 0.0012–0.0046); meanwhile, the mean leave-one-condition-out accuracy gap between high- and low-stability quartiles (5.2 percentage points) closely matches that of the default contiguous partitioning (5.1 percentage points). Furthermore, as the number of blocks varies across $B \in \{16, 32, 128, 256\}$, the item-level correlation remains between 0.917 and 0.983 (reaching 0.971, 0.983, and 0.970 for $B=32$, $128$, and $256$, respectively), the model-level ranking correlation reaches 0.939–0.986, the AUROC for discriminating semantic variants from hard negatives reaches 0.901–0.958 ($B \in [32, 256]$), and representative models such as DeepSeek-VL-7B and LLaVA-1.5-13B consistently maintain substantial quartile accuracy advantages of 13.5–16.0 percentage points. Therefore, KSS reliably characterizes distributed internal knowledge support, demonstrating that its evaluation effectiveness neither relies on a specific index ordering of hidden dimensions nor suffers from sensitivity to channel-block hyperparameters.

\subsection{Entity-Recognition Conditioned Subset Analysis}
To verify that prediction flips and internal stability degradation under natural viewpoint variations stem from unstable latent knowledge retrieval rather than upstream visual entity misrecognition, we design an explicit visual entity identification filtering experiment using same-category hard distractors (species within the same taxonomic order or landmarks of the same architectural type) under two option orders, isolating a strictly controlled, recognition-verified subset in which the model accurately identifies the entity across the reference, natural close-view, and natural far-view images. The results show that even when the model unambiguously recognizes the target entity across all viewpoints, cross-view factual answer flips and internal support degradation persist, and both MRS and KSS retain strong, statistically significant predictive power for cross-condition accuracy. Specifically, across Qwen3.5-9B, InternVL3.5-38B, DeepSeek-VL-7B, and LLaVA-1.5-13B, the close- and far-view answer flip rates (3.0\%–6.6\% and 3.2\%–6.7\%, respectively), mean MRS (0.906–0.978), and mean KSS (0.882–0.952) on the recognition-verified subset remain highly consistent with those on the unfiltered probed set (flip rates of 3.1\%–6.5\% and 3.4\%–7.2\%, mean MRS of 0.896–0.978, and mean KSS of 0.876–0.950). Under the leave-one-condition-out evaluation protocol, the accuracy gaps between high- and low-MRS quartiles on this recognition-verified subset are significantly positive across all four models, reaching 19.4 percentage points on DeepSeek-VL-7B (95\% CI $[15.7, 22.5]$; logistic regression odds ratio $\text{OR} = 1.65$ after controlling for reference confidence, $p < 10^{-4}$), 10.7 percentage points on LLaVA-1.5-13B (95\% CI $[9.0, 12.3]$, $\text{OR} = 1.38$, $p < 10^{-4}$), 5.2 percentage points on Qwen3.5-9B (95\% CI $[3.9, 6.5]$, $\text{OR} = 1.17$, $p < 10^{-4}$), and 2.6 percentage points on InternVL3.5-38B (95\% CI $[0.3, 4.8]$); meanwhile, KSS achieves substantial quartile accuracy gains of 14.5 percentage points on DeepSeek-VL-7B (95\% CI $[11.4, 17.4]$, $\text{OR} = 1.52$, $p < 10^{-4}$) and 11.0 percentage points on LLaVA-1.5-13B (95\% CI $[9.5, 12.7]$, $\text{OR} = 1.40$, $p < 10^{-4}$) on this controlled subset, even surpassing the unfiltered baselines (13.4 and 12.1 percentage points, with ORs of 1.47 and 1.34, respectively). Therefore, accuracy degradation and internal representational shifts across viewpoints cannot be attributed to upstream visual perception failures, confirming that MRS and KSS capture the stability of latent factual knowledge retrieval and support beyond visual entity recognition.

\subsection{Decomposition of Support Magnitude and Redistribution}
To disentangle the respective contributions of total support magnitude and internal support redistribution in latent knowledge support-structure stability (KSS), we conduct an orthogonal decomposition experiment that factorizes layer-wise KSS into pure magnitude retention $\text{KSS}_{\text{mag}}^l = \frac{2\min(M_a, M_t)}{M_a + M_t}$ and normalized structural overlap $\text{KSS}_{\text{dist}}^l = \sum_b \min(a_b/M_a, t_b/M_t)$, evaluating answer-flip prediction AUROC across both the four representative models (196,464 pairs) and all 15 MLLMs (924,528 reference-correct pairs) using 5-fold item-grouped cross-validated logistic regression controlling for reference margin, model, condition, and question template. The results show that KSS variations are predominantly governed by structural redistribution $\text{KSS}_{\text{dist}}$, both components independently and significantly boost answer-flip discrimination over the baseline, and $\text{KSS}_{\text{dist}}$ provides statistically significant incremental predictive power beyond $\text{KSS}_{\text{mag}}$. Specifically, $\text{KSS}_{\text{dist}}$ exhibits a high Spearman rank correlation of 0.965--0.995 with full KSS (compared with 0.70--0.92 for $\text{KSS}_{\text{mag}}$); relative to the reference-margin baseline (0.835 and 0.826 AUROC on the 4-model and 15-model sets, respectively), full $\text{KSS}$, $\text{KSS}_{\text{mag}}$, and pure redistribution $\text{KSS}_{\text{dist}}$ yield substantial AUROC gains of $+5.13$ (95\% CI $[4.71, 5.54]$), $+5.03$ ($[4.60, 5.46]$), and $+4.79$ ($[4.36, 5.19]$) percentage points on the four representative models, and consistently achieve gains of $+4.56$ ($[4.48, 4.79]$), $+4.84$ ($[4.78, 5.06]$), and $+4.22$ ($[4.13, 4.44]$) percentage points across all 15 models. Crucially, in nested comparisons, adding the structural redistribution component $\text{KSS}_{\text{dist}}$ on top of $\text{KSS}_{\text{mag}}$ yields statistically significant incremental AUROC gains of $+0.44$ percentage points (95\% CI $[0.35, 0.55]$) on the four representative models and $+0.29$ percentage points (95\% CI $[0.25, 0.32]$) across all 15 models (with joint $\text{KSS}_{\text{mag}} + \text{KSS}_{\text{dist}}$ reaching $+5.47$ and $+5.12$ percentage points, respectively); moreover, incorporating KSS on top of MRS ($\text{MRS} + \text{KSS}$) further improves AUROC by $+0.47$ percentage points (95\% CI $[0.39, 0.54]$, reaching $+6.90$ percentage points over baseline) on the four models and $+0.07$ percentage points (95\% CI $[0.07, 0.08]$, reaching $+7.96$ percentage points over baseline) across all 15 models, with $\text{KSS}_{\text{dist}}$ retaining a significant positive increment of $+0.06$ percentage points (95\% CI $[0.01, 0.13]$) even after jointly controlling for $\text{MRS}$ and $\text{KSS}_{\text{mag}}$ on the four models. Therefore, KSS does not merely reflect global support-magnitude scaling; rather, internal support redistribution ($\text{KSS}_{\text{dist}}$) carries essential vulnerability signals beyond support magnitude and output confidence, validating the joint formulation of magnitude and structural consistency in KSS.

\subsection{Human Validation and Quality Assurance}
To ensure the reliability and quality of MMStableBench, we conducted an additional human validation process involving five researchers from academic institutions, all holding Ph.D. degrees. These researchers independently inspected the benchmark data, focusing on factual correctness, semantic consistency between questions and answers, and the validity of knowledge-preserving variations across different elicitation conditions. This human verification process was designed to identify potential annotation errors, ambiguous questions, and inconsistencies introduced during automated data construction, thereby strengthening the overall quality and reliability of MMStableBench.

\subsection{Additional Qualitative Examples under Visual Variations}
\label{app:additional-visual-analysis}

\begin{figure*}[t]
\centering
\includegraphics[width=\linewidth]{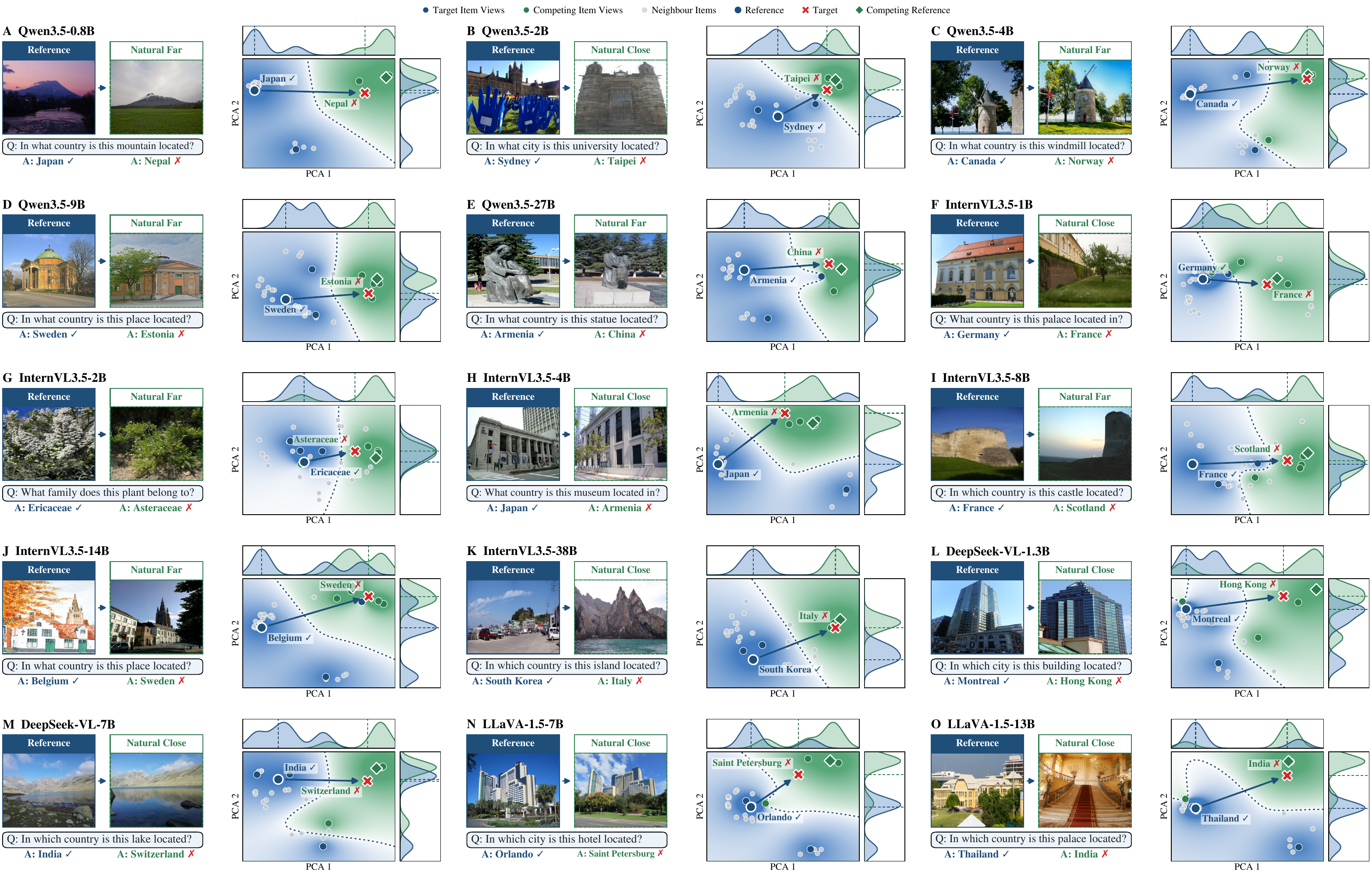}
\vspace{-0.3cm}
\caption{\textbf{Additional Qualitative Examples of MRS under Visual Variations.}}
\vspace{-0.4cm}
\label{fig:fig_more_mrs}
\end{figure*}

\noindent\textbf{MRS visualizations across models.}
Figure~\ref{fig:fig_more_mrs} extends the visual-variation analysis in the main paper with examples from all 15 evaluated MLLMs. In each example, a natural close- or far-view image of the same entity is paired with the unchanged question and answer options. Although the model answers the reference input correctly, the alternative view leads to an incorrect prediction. The PCA visualizations place the reference and target views alongside neighboring and competing-item views. Across model families, the target representations shift toward regions associated with competing answers, illustrating how a knowledge-preserving viewpoint change can alter the model's reference-answer preference. These are selected qualitative cases rather than estimates of the frequency of such failures.

\begin{figure*}[t]
\centering
\includegraphics[width=\linewidth]{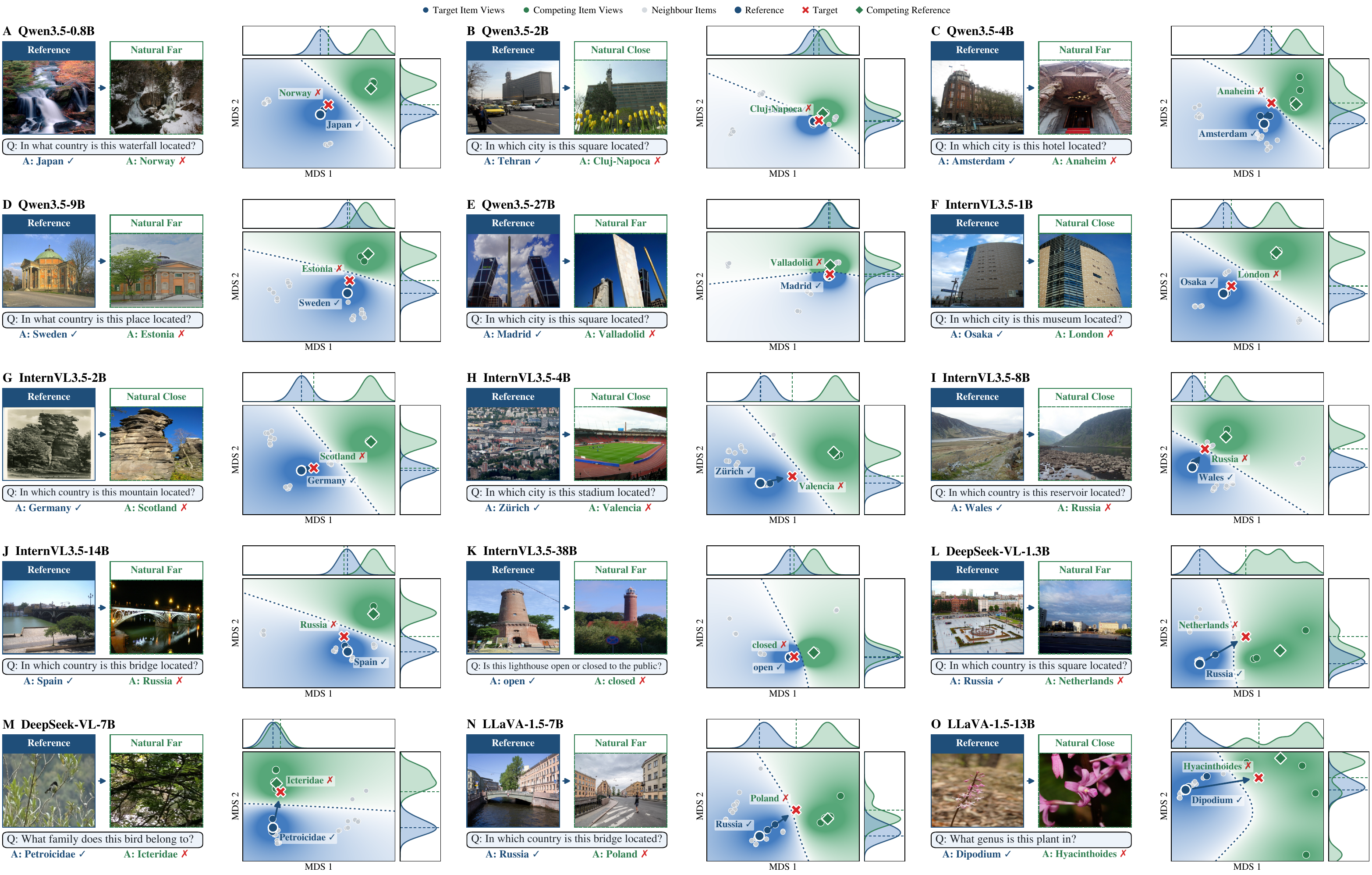}
\vspace{-0.3cm}
\caption{\textbf{Additional Qualitative Examples of KSS under Visual Variations.}}
\vspace{-0.4cm}
\label{fig:fig_more_kss}
\end{figure*}

\noindent\textbf{KSS visualizations across models.}
Figure~\ref{fig:fig_more_kss} provides complementary MDS visualizations of reference-correct items whose predictions change under natural image variations. Although the question and underlying knowledge are unchanged, the target views occupy different regions from the reference views in the projected representations, often closer to competing-item views. These examples illustrate that visually induced changes in model behavior can be accompanied by changes in the organization of internal knowledge support. The projections provide qualitative context for KSS, while KSS itself is computed from the reference-gradient support profiles rather than from distances in the two-dimensional plots.

\begin{figure*}[t]
\centering
\includegraphics[width=\linewidth]{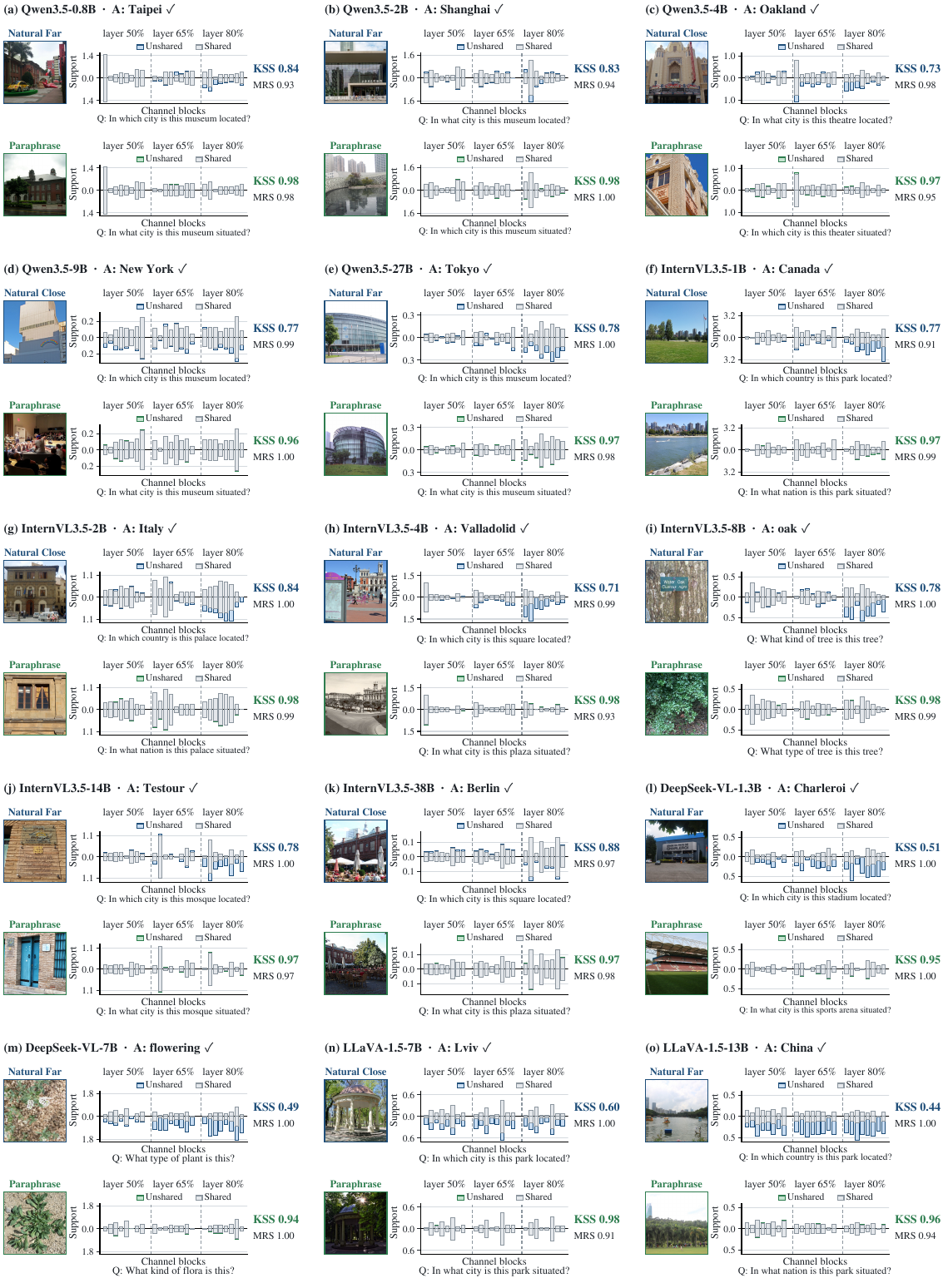}
\vspace{-0.3cm}
\caption{\textbf{Additional Correctness-Preserving Cases across 15 MLLMs.}}
\vspace{-0.4cm}
\label{fig:fig_more_results}
\end{figure*}

\begin{figure*}[t]
\centering
\includegraphics[width=\linewidth]{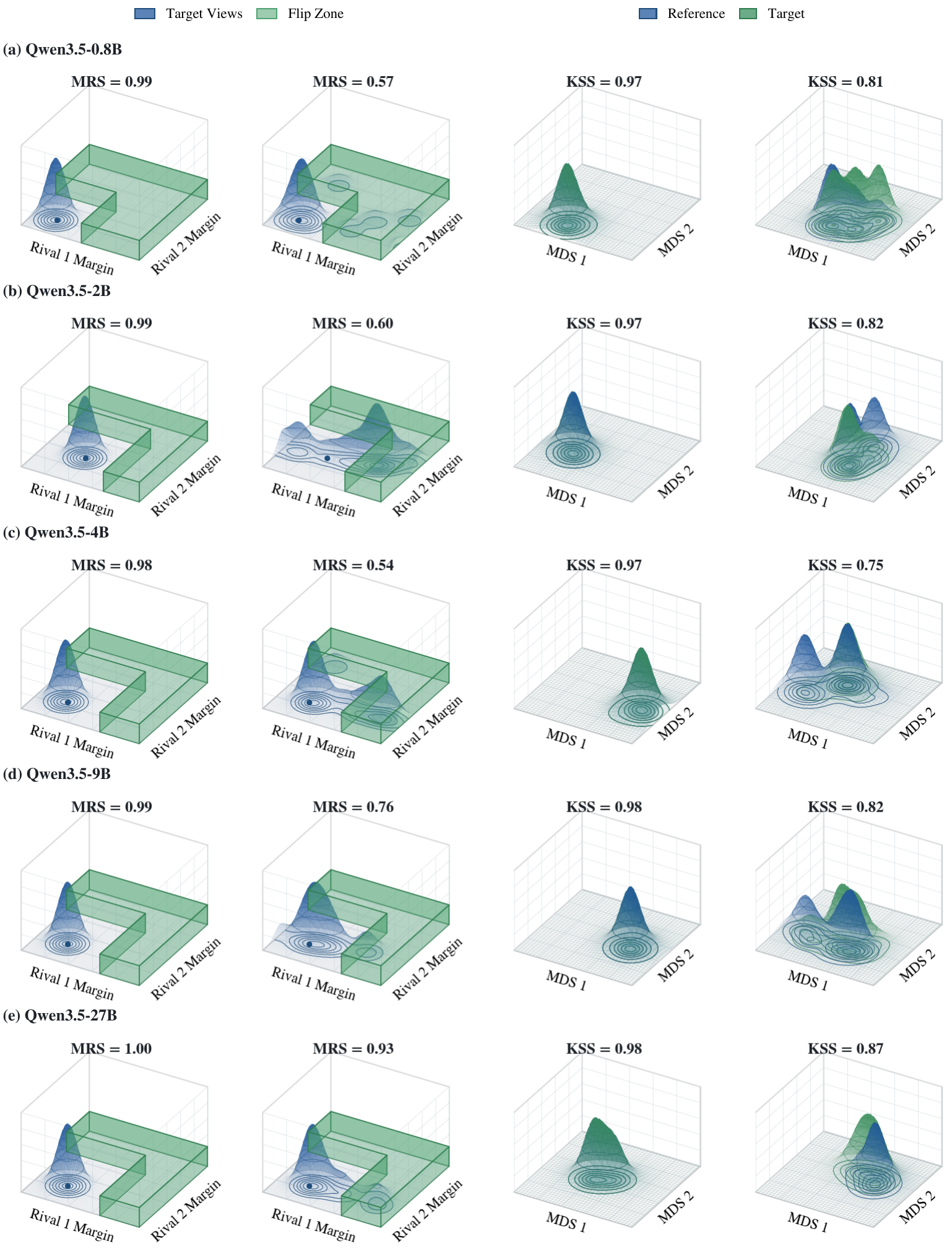}
\vspace{-0.3cm}
\caption{\textbf{Geometric Interpretation of MRS and KSS: Qwen3.5 Models.}}
\vspace{-0.4cm}
\label{fig:fig7_more_results_p1}
\end{figure*}

\begin{figure*}[t]
\centering
\includegraphics[width=\linewidth]{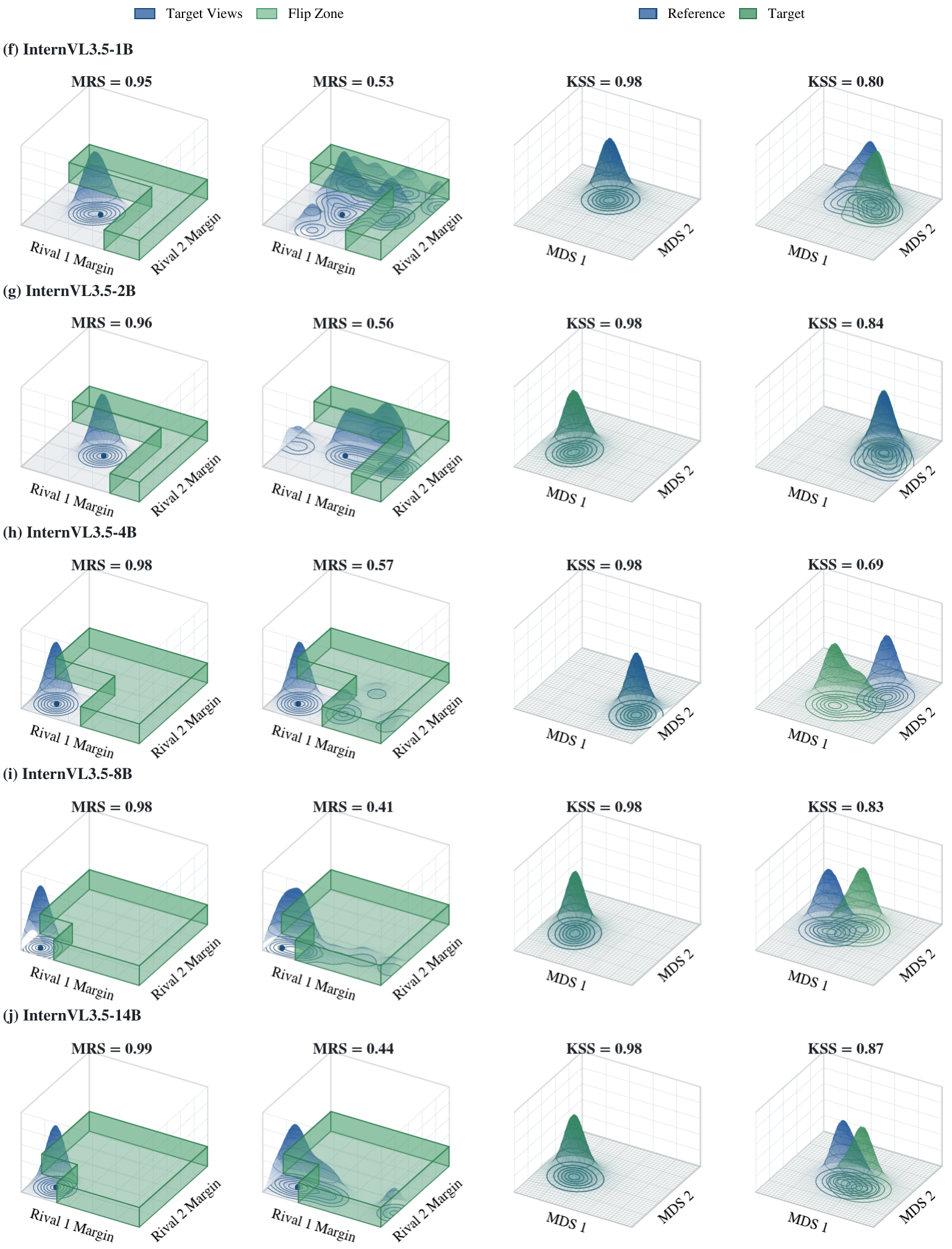}
\vspace{-0.3cm}
\caption{\textbf{Geometric Interpretation of MRS and KSS: InternVL3.5 Models (I).}}
\vspace{-0.4cm}
\label{fig:fig7_more_results_p2}
\end{figure*}

\begin{figure*}[t]
\centering
\includegraphics[width=\linewidth]{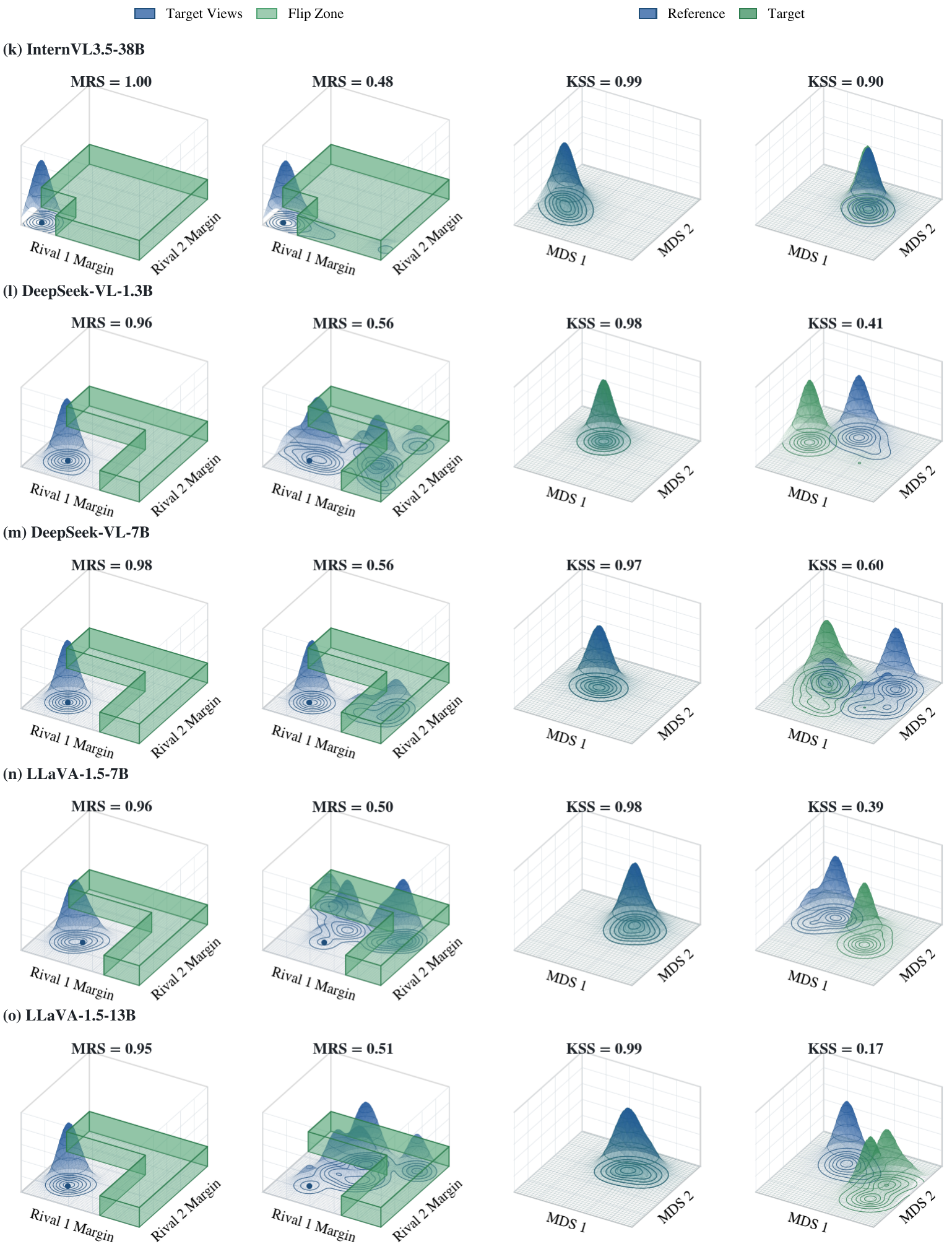}
\vspace{-0.3cm}
\caption{\textbf{Geometric Interpretation of MRS and KSS: Remaining Models.}}
\vspace{-0.4cm}
\label{fig:fig7_more_results_p3}
\end{figure*}

\subsection{Additional Correctness-Preserving Examples}
\label{app:additional-correctness-preserving}

\noindent\textbf{Similar answer preference, different internal support.}
Figure~\ref{fig:fig_more_results} extends the correctness-preserving qualitative analysis to all 15 models. For each model, we compare a natural-view variant and a question paraphrase of a reference-correct knowledge item. The answer remains correct in both conditions, and MRS is generally high, yet the natural-view variant has substantially lower KSS than the paraphrase. For example, DeepSeek-VL-7B has MRS of 1.00 in both conditions, while KSS differs markedly between the natural view (0.49) and the paraphrase (0.94). For LLaVA-1.5-13B, KSS is 0.44 under the natural far view versus 0.96 under paraphrasing, even though the natural-view MRS remains 1.00. The channel-block visualizations show corresponding differences in shared and unshared reference-gradient support across decoder depths. These cases illustrate that preserving the correct answer and its margin does not require preserving the same internal support profile.

\subsection{Additional Geometric Interpretations of MRS and KSS}
\label{app:additional-geometric-analysis}

\noindent\textbf{Qwen3.5 models.}
Figure~\ref{fig:fig7_more_results_p1} extends the geometric interpretation in the main paper to five Qwen3.5 models. In the two rival-margin coordinates, examples with higher MRS (0.98--1.00) remain concentrated away from the illustrated answer-flip zone, whereas examples with lower MRS (0.54--0.93) extend further toward that zone. In the MDS visualizations, higher-KSS cases (0.97--0.98) show closely aligned reference and target distributions, while lower-KSS cases (0.75--0.87) show greater differences in their location or shape. The degree of separation varies across model sizes; these examples are intended to illustrate the metrics rather than establish a monotonic scaling trend.

\noindent\textbf{InternVL3.5 models.}
Figure~\ref{fig:fig7_more_results_p2} shows the corresponding patterns for InternVL3.5 models from 1B to 14B. Examples with lower MRS exhibit greater movement of the displayed target-view distribution toward the flip zone, while lower-KSS examples show a larger mismatch between the reference and target distributions. For InternVL3.5-8B, MRS changes from 0.98 in the higher-stability case to 0.41 in the lower-stability case, whereas KSS changes from 0.98 to 0.83 in the two KSS examples. The differences illustrate why margin retention and support-structure consistency provide distinct perspectives on stability, rather than interchangeable measurements.

\noindent\textbf{Remaining model families.}
Figure~\ref{fig:fig7_more_results_p3} includes InternVL3.5-38B, both DeepSeek-VL models, and both LLaVA-1.5 models. The separation between higher- and lower-stability cases is visible across these architectures, but its extent differs between MRS and KSS. For example, InternVL3.5-38B has a lower-MRS example of 0.48 while its lower-KSS example remains at 0.90. By contrast, LLaVA-1.5-13B shows a pronounced difference between the reference and target support distributions, with KSS decreasing from 0.99 in the higher-stability example to 0.17 in the lower-stability example. Taken together, Figures~\ref{fig:fig7_more_results_p1}--\ref{fig:fig7_more_results_p3} reinforce the complementary interpretations of MRS as resistance to reference-margin degradation and KSS as preservation of internal support structure.

\section{Limitations}
\label{lim}
Our current evaluation focuses on single-turn knowledge questions under controlled visual and linguistic variations. An interesting direction for future work is to extend this analysis to multi-turn multimodal interactions, where the same knowledge may be elicited through a sequence of related questions. Such an extension could provide additional insights into how latent knowledge stability manifests in more interactive settings.

\section{More Discussion}
\label{dis}

\noindent\textbf{Q1. Why Does MRS Use a First-Order Approximation and the Last Prompt Token?}

MRS evaluates how representational changes affect the reference answer preference through a first-order approximation, $\delta_{m,i,t}^{\ell}=J_{m,i,a}^{\ell}\Delta h_{m,i,t}^{\ell}$, where the Jacobian is fixed at the reference condition~\cite{kramar2024atp}. This formulation focuses on decision-relevant changes rather than arbitrary representational drift. We use the last prompt token because its residual state integrates preceding visual and textual information through causal attention~\cite{zhang2025cross}, providing a consistent representation for evaluating changes across elicitation conditions.

MRS is inherently local and one-sided. To verify the first-order approximation, we compare it with single-site activation replacement and with the actual target-condition forward pass. At the 80\% layer, the approximation closely matches activation replacement (Spearman 0.88--1.00 across six models). Across all 15 models, the layer-averaged estimate agrees with the actual margin change (average Spearman 0.82) and identifies margin decreases consistently in 85\% of cases, with no degradation for larger displacements. Changes that strengthen the reference answer preference are not penalized; thus, MRS measures resistance to reference-margin degradation, while KSS complements it by capturing changes in internal support even when answer preferences remain stable.

\noindent$\triangleright$ \textbf{\textit{Q2. How Is Knowledge Stability Distinguished from Visual Recognition Stability?}}

A key distinction is that visual recognition stability concerns whether an MLLM consistently identifies the same entity across different visual inputs~\cite{fan2025v2bench}, whereas latent knowledge stability concerns whether previously acquired knowledge can be consistently accessed and utilized across different elicitation conditions~\cite{chou2025mm}. Although both processes contribute to multimodal question answering, they characterize different aspects of model behavior. MMStableBench focuses on the latter by preserving the underlying entity--question--answer fact across visual and linguistic variations, with text-only probes establishing that the evaluated knowledge is already accessible to the model. Rather than measuring entity recognition accuracy, MRS evaluates the retention of answer preferences, while KSS characterizes the consistency of internal knowledge support. Furthermore, linguistic variations and answer-option permutations preserve the visual input, allowing stability to be examined without changing the visual evidence. Correctness-preserving cases further demonstrate that internal knowledge support can vary substantially even when the model consistently produces the correct answer. Together, these analyses emphasize that our evaluation goes beyond visual recognition robustness to characterize how consistently MLLMs access and utilize previously demonstrated knowledge across multimodal elicitation conditions.

\noindent\textbf{$\triangleright$ Q3. What Does Internal Knowledge Support Mean in KSS, and What Is Its Independent Predictive Value?}

A key distinction is that in MLLMs, knowledge-related decisions emerge from distributed representations rather than isolated single neurons [7, 8]. KSS characterizes this distributed support through reference-gradient-weighted block activations $d_{b}^{l}(x)$, which formulate the first-order attribution patching approximation to the causal effect of jointly ablating channel block $b$ on the verified-answer margin (validated against activation replacement in Q1). By fixing the reference gradient across conditions, KSS evaluates whether the same functional components continue to drive the verified answer. Crucially, KSS provides independent predictive power where output confidence and MRS saturate: because MRS measures net margin change, losing support in original channels can be masked by compensatory positive activations in other channels ($\text{MRS} \approx 1.00$), whereas KSS's weighted Dice overlap detects this internal redistribution (with 12.4\% of top-MRS-quartile correct inputs falling into the bottom KSS quartile). Consequently, under strictly matched output confidence, low-KSS items drop from 96.9\% to 89.3\% accuracy across $k=16$ repeated perturbations while high-KSS items remain at 97.7\%. KSS also independently anticipates held-out condition accuracy under leave-one-condition-out evaluation (up to a 15.6\% gap) and boosts answer-flip AUROC over the reference-margin baseline to 88.1\%, surpassing raw drift (85.4\%) and mismatched-gradient controls ($B_4$, 85.3\%).

\noindent$\triangleright$ \textbf{\textit{Q4. How Are Answer Choices Aligned under Option Permutation?}}

A key consideration is that option permutation reassigns answer labels to different semantic choices, so reference and target inputs must be compared by semantic choice identity rather than label position. The reference Jacobian $J_{m,i,a}^{\ell}$ and margin gradient $\boldsymbol\gamma^{\ell}_{m,i}$ are computed once on the reference input, whose choice order defines the semantic choice order of each knowledge item, and are held fixed across all target conditions. Target conditions enter MRS and KSS only through their hidden states, as representational displacements for MRS and gradient-weighted block contributions for KSS, so the Jacobian and gradient require no relabeling and all target representations are evaluated under the same reference decision directions. The label-to-choice mapping of each permuted item is applied only to the target answer logits, which are reordered by semantic choice identity before computing accuracy and answer flips. Under this design, option permutation probes whether the internal decision readout remains invariant when the same knowledge is presented under a different answer-label assignment. A model may therefore preserve its final answer while its internal readout shifts with the label assignment, which is consistent with the larger fraction of cross-condition drift lying within the decision subspace under option permutation (10.1\%) than under paraphrasing and natural-view variations (2.2\%). This sensitivity is complementary to the output-level accuracy and is precisely what stability evaluation beyond output correctness is intended to reveal.

\noindent$\triangleright$ \textbf{\textit{Q5. Are KSS Comparisons across Models Affected by Different Reference-Correct Subsets?}}

A key consideration is that KSS is computed on each model's reference-correct items, so different models are evaluated on different item subsets, ranging from 12{,}817 items for InternVL3.5-1B to 17{,}385 items for Qwen3.5-27B. To separate model stability from differences in item composition, we recompute both metrics on the 6{,}372 items answered correctly under the reference condition by all 15 models. Model rankings remain highly consistent with those in Table~1 in the main paper (Spearman $\rho=0.982$ for KSS and $0.968$ for MRS), and all scaling trends discussed in the main text are preserved. For example, as InternVL3.5 scales from 1B to 38B, KSS changes from 0.914 to 0.932 with non-monotonic variations, and scaling Qwen3.5 from 9B to 27B still decreases MRS while increasing KSS. Within-family orderings are also unchanged, although the KSS gaps between the smaller and larger DeepSeek-VL and LLaVA-1.5 models narrow on the common subset, suggesting that part of these gaps reflects additional items that larger models answer correctly but support less stably. For the high-MRS/low-KSS statistic in the main text, quartiles are computed within each model over inputs answered correctly under both the reference and target conditions, and the resulting counts are pooled across models. In this way, the reported cross-model comparisons reflect differences in latent knowledge stability rather than differences in the evaluated item subsets.

\noindent\textbf{$\triangleright$ Q6. Does Holding Reference Directions Fixed under Option Permutation Introduce a Confounding Bias?}

A natural question is whether fixing the reference Jacobian $J_{m,i,a}^{l}$ and margin gradient $\gamma_{m,i}^{l}$ under option permutation penalizes models simply for shifting their output token toward the relocated answer letter. Both mechanistic properties and empirical controls confirm that no such confounding bias undermines our evaluation. First, MRS and KSS are evaluated at intermediate decoder depths ($\mathcal{L}_m \in \{50\%, 65\%, 80\%\}$) rather than at the final token-unembedding layer. Because the remaining 20\%--50\% of decoder layers attend to the prompt options to perform late-stage symbol binding, backpropagating the reference margin to $\mathcal{L}_m$ yields directions dominated by upstream semantic content rather than literal option-letter identity. Second, if fixed reference directions mathematically penalized accurate label reassignment, models with high option-permutation accuracy would systematically exhibit low MRS and KSS. Instead, Table~1 demonstrates that capable models achieve near-optimal scores under option permutation (e.g., 0.971 MRS and 0.954 KSS for InternVL3.5-38B; 0.970 MRS and 0.914 KSS for Qwen3.5-9B), proving that intermediate representations can remain highly stable under fixed reference directions while the final readout adapts to the permuted label. Third, our experimental conclusions do not rely on option permutation: in Figure~5, stability measured exclusively on the four label-preserving conditions predicts held-out accuracy under option permutation, and in Table~4 (Drop Opt.\ Perm.), excluding option permutation entirely preserves model-, item-, and relation-level correlations of 0.94/0.93/0.93 for MRS and 0.96/0.92/0.87 for KSS.

\end{document}